\documentclass{article}

\usepackage{microtype}
\usepackage{graphicx}
\usepackage{subcaption}
\usepackage{booktabs} 

\usepackage{hyperref}

\usepackage[preprint]{icml2027}

\usepackage{amsmath}
\usepackage{amssymb}
\usepackage{mathtools}
\usepackage{amsthm}

\usepackage{amsmath}
\usepackage{amssymb}
\usepackage{mathtools}
\usepackage{amsthm}

\usepackage[capitalize,noabbrev]{cleveref}

\usepackage[utf8]{inputenc} 
\usepackage[T1]{fontenc}    
\usepackage{url}            
\usepackage{booktabs}       
\usepackage{amsfonts}       
\usepackage{nicefrac}       
\usepackage{microtype}      
\usepackage{xcolor}         
\usepackage{makecell}
\usepackage{graphicx}
\usepackage{amsmath}
\usepackage{multirow}
\usepackage{wrapfig}
\usepackage{enumitem}
\usepackage{wrapfig}
\usepackage{float}
\usepackage{graphicx}
\usepackage{caption}
\usepackage{color}
\usepackage{colortbl}
\usepackage{xcolor}
\usepackage{bbding}
\usepackage{pifont}
\usepackage{tabularx} 
\usepackage{xspace}          
\usepackage{booktabs}
\usepackage{algorithmic}
\usepackage{algorithm}
\usepackage{natbib}   

\usepackage[most]{tcolorbox}
\usepackage{xcolor}
\usepackage{fontawesome5}
\usepackage{wrapfig}
\usepackage{tabularx}
\usepackage{adjustbox}
\usepackage{array}
\usepackage{xcolor}
\usepackage{graphicx}

\theoremstyle{plain}

\theoremstyle{definition}

\theoremstyle{remark}

\newcommand{\ourmethod}{\textbf{FlashMorph2}}

\usepackage[textsize=tiny]{todonotes}

\icmltitlerunning{On-Policy Morphing into Hybrid Attention Models}

\begin{document}

\twocolumn[
  \icmltitle{On-Policy Morphing into Hybrid Attention Models}



  \icmlsetsymbol{equal}{*}

  \begin{icmlauthorlist}
    \icmlauthor{Disen Lan}{fdu,seed,equal}
    \icmlauthor{Xuanda Wang}{seed}
    \icmlauthor{Jianbin Zheng}{seed}
    \icmlauthor{Huafeng Kuang}{seed}
    \icmlauthor{Wei Tang}{seed}
    \icmlauthor{Yu Cheng}{ntu}
  \end{icmlauthorlist}

  \icmlaffiliation{fdu}{Fudan University}
  \icmlaffiliation{seed}{ByteDance Seed}
  \icmlaffiliation{ntu}{Nanyang Technological University}

  \icmlcorrespondingauthor{Yu Cheng}{yu.cheng@ntu.edu.sg}

  \icmlkeywords{Machine Learning, ICML}

  \vskip 0.3in
]



\printAffiliationsAndNotice{\frenchspacing\textsuperscript{*}Work done at ByteDance Seed}  

\begin{abstract}
  Hybrid attention models combine full and linear attention to achieve a better quality--efficiency trade-off. Converting pretrained Transformers into such models requires identifying which layers should retain full attention under a prescribed budget. We identify two \textit{off-policy} mismatches in FlashMorph's layer selection: (i) continuous branch interpolation and teacher-forced hidden states differ from discrete hybrid computation; and (ii) frozen model weights fail to account for parameter adaptation during subsequent recovery. We introduce \textbf{FlashMorph2}, an \textit{on-policy} layer-selection method that (I) evaluates sampled discrete hybrid architectures end to end using their own hidden states, and (II) jointly optimizes layer-selection scores and model weights under the recovery objective to reflect post-adaptation performance. Experiments across Qwen3 model scales and linear-attention variants show that FlashMorph2 consistently outperforms existing layer-selection methods, improving average scores over FlashMorph by $+2.42$ points on RULER, $+1.08$ on LongBench, and $+0.76$ on recall-intensive tasks, while completing layer selection in only $0.72$--$3.57$ GPU hours across 0.6B--8B models, demonstrating high efficiency and scalability.
\end{abstract}

\section{Introduction}

Hybrid attention models retain full attention in only a few layers and use linear attention elsewhere, preserving long-context modeling capabilities and precise retrieval~\citep{arora2024simple,jelassi2024repeat,li2026distilling}, while substantially reducing computation and KV-cache memory usage~\citep{katharopoulos2020transformers,sun2025linear-moe,sun2025survey}. Owing to this favorable trade-off between performance and efficiency, hybrid attention has been adopted in a growing number of production-scale large language models (LLMs)~\citep{li2025minimax,team2025kimi,cao2026qwen3,qwen35blog,team2026kimi}. To avoid the prohibitive cost of pretraining from scratch, existing approaches reuse the weights of pretrained LLMs and replace a subset of full attention layers with linear attention, achieving Transformers-to-hybrid conversion. Model quality is then recovered through distillation and continued training~\citep{kasai2021finetuning,bick2024transformers,lan2025liger,goldstein2025radlads,wang2024mamba,bick2025llamba,zhang2025lolcats}. A central challenge in this conversion process is determining which full attention layers to retain under a given budget. As the only layers in the resulting hybrid model that can directly access the entire context, their placement largely determines how much of the original model's capability can be recovered~\citep{li2026distilling,chen2026hybrid,lan2026morphing}.

\begin{figure}[!t]
    \centering
    \includegraphics[width=\linewidth]{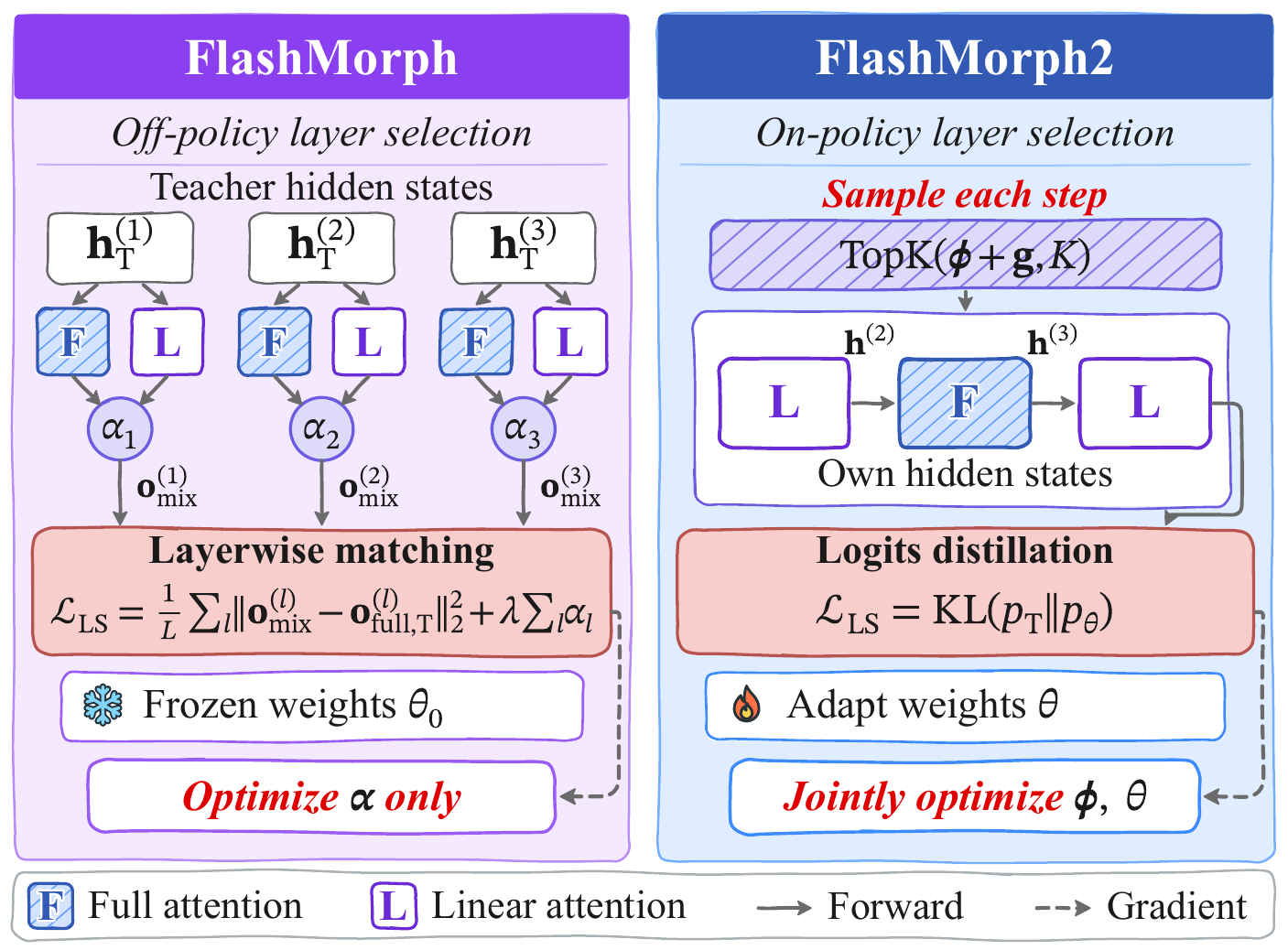}
    \vspace{-1.2em}
    \caption{\textbf{Comparison between FlashMorph and FlashMorph2.} FlashMorph uses \textit{off-policy} teacher-forced continuous gates with frozen model weights, whereas FlashMorph2 samples \textit{on-policy} discrete hybrid architectures and jointly optimizes layer-selection scores and model weights.}
    \label{fig:flashmorph_compare}
    \vspace{-1.2em}
\end{figure}

Exhaustively evaluating all possible layer combinations is prohibitively expensive, motivating existing work to explore a variety of layer-selection strategies. Beyond fixed uniform interleaving patterns, PostNAS~\citep{gu2026jet} performs supernet-based architecture search, KL-LS~\citep{li2026distilling} ranks layers by the KL reduction from per-layer restoration and distillation, and HALO~\citep{chen2026hybrid} uses replacement-based sensitivity scoring. More recently, FlashMorph~\citep{lan2026morphing} jointly optimizes layer-wise gates to learn a global hybrid configuration. Each layer interpolates between aligned full- and linear-attention branches through a continuous gate, which is then discretized under a given full-attention budget. FlashMorph preserves strong long-context recall and general benchmark performance while substantially reducing layer-selection cost.

\textbf{Limitations.} FlashMorph's layer-selection process still suffers from two forms of \textit{off-policy} mismatch. \textbf{(L1)} \textit{\textbf{Off-policy computation}}: The forward computation used during layer selection differs from that of the candidate discrete model. Continuous mixing allows the full- and linear-attention branches to complement each other, while each layer operates on teacher-forced hidden states. At deployment, only one branch is retained per layer and hidden states are generated by the hybrid model itself. Consequently, complementarities observed under continuous mixing may not transfer to the final discrete architecture, and the resulting layer-importance estimates may fail to capture error accumulation and cross-layer compensation. \textbf{(L2)} \textit{\textbf{Off-policy weights}}: The weights used for layer selection are inherited from the original model and remain frozen during gate optimization, without adapting to the hybrid architecture induced by the selection policy. Layer selection therefore cannot account for the cross-layer co-adaptation of model weights during subsequent recovery, which may alter the layer-importance ranking observed before recovery.

\textbf{Our Method.} We introduce \ourmethod, which improves upon FlashMorph by bringing the \textit{on-policy} principle into its layer-selection process. To address \textbf{(L1)}, FlashMorph2 samples a discrete hybrid attention architecture that satisfies the prescribed full-attention budget and performs an end-to-end forward pass through the sampled architecture, allowing each layer to operate on hidden states produced by the model itself. To address \textbf{(L2)}, FlashMorph2 jointly trains model weights across these sampled hybrid attention architectures using the same optimization objective as in subsequent recovery training, allowing layer selection to account for performance after weight adaptation.

\textbf{Results.} Across Qwen3-0.6B and Qwen3-1.7B backbones and three linear-attention variants, FlashMorph2 consistently outperforms all layer-selection baselines, improving average scores over FlashMorph by $+2.42$ points on RULER, $+1.08$ on LongBench, and $+0.76$ on recall-intensive tasks while preserving language modeling and commonsense reasoning capabilities. Moreover, FlashMorph2 completes layer selection in only $0.72$--$3.57$ GPU hours across 0.6B--8B models, running $8.4$--$20.8\times$ faster than HALO and $560$--$854\times$ faster than KL-LS.

\textbf{Contributions.} Our key contributions are as follows:

\begin{itemize}[leftmargin=1.6em, topsep=-1.9pt, itemsep=1pt, parsep=0pt, partopsep=0pt]
    \item We identify two forms of \textit{off-policy} mismatch in FlashMorph's layer selection: (I) continuous interpolation and teacher-forced hidden states differ from deployed discrete hybrid computation, and (II) scores computed with frozen weights fail to reflect the optimal layer ranking after recovery training.
    \item  We introduce FlashMorph2, an \textit{on-policy} layer selection method for Transformer-to-hybrid conversion that performs end-to-end forward passes through sampled hybrid architectures under a full-attention budget, allowing selection to account for self-generated hidden states and weights adapted under the recovery objective.
    \item We evaluate FlashMorph2 across multiple Qwen3 model scales and linear-attention variants, demonstrating consistent improvements on long-context and recall-intensive benchmarks while maintaining high layer-selection efficiency and scalability.
\end{itemize}
\section{Related Work}

\textbf{Layer Selection for Transformer-to-hybrid Conversion.} Transformer-to-hybrid conversion requires selecting which full-attention layers to retain under a prescribed budget. A simple solution is fixed uniform interleaving~\citep{lan2025liger,goldstein2025radlads}, but it ignores layer-wise heterogeneity in pretrained models. PostNAS~\citep{gu2026jet} trains a once-for-all supernet and uses task-guided beam search to determine full-attention layer placement, incurring substantial training and search costs; KL-LS~\citep{li2026distilling} ranks layers by post-distillation KL loss after single-layer restoration; and HALO~\citep{chen2026hybrid} ranks full-attention layers by the downstream performance changes induced by their individual replacement with linear-attention layers. These layerwise scoring methods incur substantial selection overhead from repeated single-layer evaluations and fail to account for interactions among layers in the resulting hybrid architecture. FlashMorph~\citep{lan2026morphing} jointly optimizes layerwise gate values, yielding more effective hybrid configurations at substantially lower selection cost. However, these methods remain \textit{off-policy} due to mismatches in forward computation or model weights between the selection-time and recovered hybrid models. FlashMorph2 improves upon FlashMorph with \textit{on-policy} layer selection that performs end-to-end forward passes through sampled discrete hybrid architectures while accounting for the effects of weight adaptation during recovery.

\textbf{On-Policy Learning.} Teacher forcing relies on ground-truth prefixes and is therefore \textit{off-policy} with respect to the model's own generation policy; the resulting train-inference distribution mismatch can lead to exposure bias and error accumulation~\citep{bengio2015scheduled}. To mitigate this distribution shift, \textit{on-policy} learning emphasizes obtaining supervision on states induced by the learner's own policy~\citep{ross2011reduction,lin2020autoregressive}. Following this principle, \textit{on-policy} distillation (OPD) trains students under teacher supervision on sequences generated by the students themselves~\citep{lu2025onpolicydistillation,li2026rethinking,ma2026mopd}. FlashMorph2 extends this idea to layer selection by sampling discrete hybrid architectures and estimating layer importance using self-generated hidden states and weights adapted under the recovery objective.

\section{Preliminaries}

\textbf{Notation.} Consider a pretrained Transformer~\citep{vaswani2017attention} with $L$ layers, indexed by $[L]=\{1,\dots,L\}$, and a token sequence $\mathbf{x}_{1:T}=(\mathbf{x}_1,\dots,\mathbf{x}_T)$ of length $T$. Let $\mathbf{h}^{(l)}$ denote the input hidden states to layer $l$, and let $A_{\text{full}}^{(l)}$ and $A_{\text{lin}}^{(l)}$ denote the corresponding full-attention and linear-attention modules, respectively. At the position $t$, we define the query, key and value vectors (with layer and head indices omitted) as $\mathbf{q}_t=\mathbf{h}_t\mathbf{W}_Q$, $\mathbf{k}_t=\mathbf{h}_t\mathbf{W}_K$, and $\mathbf{v}_t=\mathbf{h}_t\mathbf{W}_V$, respectively, where $\mathbf{q}_t$, $\mathbf{k}_t$, $\mathbf{v}_t \in \mathbb{R}^{1 \times d}$ and $\mathbf{W}_O$ denotes the output projection matrix.

\textbf{Full attention.} Under causal masking, full attention computes a weighted sum of value vectors in the current prefix, with weights determined by query-key similarities:
\begin{equation}
\begin{aligned}
    \mathbf{o}_{\mathrm{full},t}
    =
    \frac{
    \displaystyle\sum_{i=1}^{t}
    \exp\!\left(\mathbf{q}_t\mathbf{k}_i^\top/\sqrt{d}\right)\mathbf{v}_i
    }{
    \displaystyle\sum_{j=1}^{t}
    \exp\!\left(\mathbf{q}_t\mathbf{k}_j^\top/\sqrt{d}\right)
    }
    \mathbf{W}_O.
\end{aligned}
\end{equation}
This formulation incurs $\mathcal{O}(T^2)$ computational cost over the sequence and requires $\mathcal{O}(T)$  KV-cache memory to store past keys and values during autoregressive decoding.

\textbf{Linear attention.} Linear attention mitigates these inefficiencies by summarizing the sequence history in a fixed-size recurrent state. A broad class of linear-attention variants admits the following recurrent formulation:
\begin{equation}
\begin{aligned}
\mathbf{o}_{\text{lin},t} = \mathbf{q}_t \mathbf{S}_t \mathbf{W}_O,\quad \mathbf{S}_t=\mathbf{S}_{t-1}+\mathbf{k}_t^\top\mathbf{v}_t,
\end{aligned}
\end{equation}
where $\mathbf{S}_t \in \mathbb{R}^{d \times d}$ and $\mathbf{S}_0=0$. Different linear attention variants extend the recurrence with fixed state decay~\citep{qin2023transnormerllm,qin2024lightning}, data-dependent forget gates~\citep{yang2023gated,zhang2024gated,du2026mom}, or delta-rule updates~\citep{yang2024parallelizing,yang2025gated,hu2025improving}. These recurrences incur $\mathcal{O}(T)$ computational cost over the sequence and require $\mathcal{O}(1)$ cache memory during autoregressive decoding.

\textbf{Hybrid attention.} Hybrid attention combines full and linear attention across layers to preserve long-context retrieval capabilities while improving inference efficiency. Let $\mathcal{I} \subseteq [L]$ denote the set of layers retaining full attention; the remaining layers $[L] \setminus \mathcal{I}$ use linear attention.  We encode this hybrid configuration as a binary mask $\mathbf{m} \in \{0,1\}^L$, where $m_l=\mathbf{1} [l\in \mathcal{I}]$. The sequence-mixing output at layer $l$ is then given by
\begin{equation}
\begin{aligned}
\label{eq:hybrid_attention}
    \mathbf{o}^{(l)}
    &= m_l \mathbf{o}_{\mathrm{full}}^{(l)}
     + (1-m_l)\mathbf{o}_{\mathrm{lin}}^{(l)}, \\
    \mathbf{o}_{\mathrm{full}}^{(l)}
    &= A_{\mathrm{full}}^{(l)}\big(\mathbf{h}^{(l)}\big), \quad
    \mathbf{o}_{\mathrm{lin}}^{(l)}
    = A_{\mathrm{lin}}^{(l)}\big(\mathbf{h}^{(l)}\big).
\end{aligned}
\end{equation}
Given a budget of $K$ full-attention layers, the feasible hybrid configuration set is $\mathcal{C}_K=\{\mathcal{I}\subseteq[L]:|\mathcal{I}|=K\}$, equivalently characterized by the mask constraint $\mathbf{1}^\top \mathbf{m} = K$.  In Transformer-to-hybrid conversion, layer selection aims to identify the configuration in $\mathcal{C}_K$ that achieves the best downstream performance after recovery training.

\subsection{FlashMorph Pipeline}
\label{sec:pipeline}

FlashMorph~\citep{lan2026morphing} builds on the staged distillation pipeline of RADLADS~\citep{goldstein2025radlads} and HALO~\citep{chen2026hybrid} for Transformer-to-hybrid conversion. Treating weight transfer as initialization, we organize the subsequent pipeline into four stages: layerwise alignment (Stage 1), layer selection (Stage 2), logits distillation (Stage 3), and long-context finetuning (Stage 4).

\textbf{Stage 1: Layerwise alignment.} FlashMorph first constructs morphable layers by equipping each full-attention layer with a linear-attention branch initialized from the corresponding pretrained projection weights. The original model serves as a frozen teacher, and only the linear-attention branches are trained. Let $\mathbf{h}_{\mathrm{T}}^{(l)}$ denote the input to the teacher's attention module in layer $l$. The layerwise alignment objective is:
\begin{equation}
\begin{aligned}
\mathcal{L}_{\mathrm{align}}=\frac{1}{L}\sum_{l=1}^{L}\Big\|A_{\mathrm{lin}}^{(l)}\big(\mathbf{h}_{\mathrm{T}}^{(l)}\big)-A_{\mathrm{full}}^{(l)}\big(\mathbf{h}_{\mathrm{T}}^{(l)}\big)\Big\|_2^2
\end{aligned}
\end{equation}
Layerwise alignment trains linear-attention branches to approximate teacher attention outputs, providing alternatives to full-attention layers for subsequent layer selection and a favorable initialization for end-to-end distillation.
    
\textbf{Stage 2: Layer selection.} FlashMorph freezes the model parameters $\theta_0$ obtained after layerwise alignment and introduces a trainable continuous gate $\alpha_l \in [0, 1]$ at each layer. Both branches take teacher hidden states as input, and their outputs are interpolated as follows:
\begin{equation}
\begin{aligned}
\label{eq:flashmorph_mix}
\mathbf{o}_{\mathrm{mix}}^{(l)}=\alpha_l \mathbf{o}_{\mathrm{full},\mathrm{T}}^{(l)}+(1-\alpha_l)\mathbf{o}_{\mathrm{lin},\mathrm{T}}^{(l)}.
\end{aligned}
\end{equation}
During layer selection, only the gates are optimized by minimizing the following objective:
\begin{equation}
\begin{aligned}
\label{eq:flashmorph_stage2_loss}
\mathcal{L}_{\mathrm{LS}}=\frac{1}{L}\sum_{l=1}^{L}\Big\|\mathbf{o}_{\mathrm{mix}}^{(l)}-\mathbf{o}_{\mathrm{full},\mathrm{T}}^{(l)}\Big\|_2^2+\lambda\sum_{l=1}^{L}\alpha^{(l)},
\end{aligned}
\end{equation}
where $\lambda$ is a hyperparameter controlling the strength of linearization
regularization. After optimization, FlashMorph retains full attention in the layers indexed by $\mathcal{I}=\operatorname{TopK}(\{\alpha^{(l)}\}_{l=1}^L,K)\in\mathcal{C}_K$ and uses linear attention elsewhere. The gates are then discarded, yielding a discrete hybrid attention model for subsequent recovery training.

\textbf{Stage 3: Logits distillation.} The selected hybrid attention model $\mathcal{M}(\mathcal{I};\theta_0)$ is then distilled end to end from the frozen original Transformer, with all student parameters optimized to minimize Kullback-Leibler (KL) divergence:
\begin{equation}
\begin{aligned}
\label{eq:distill}
\mathcal{L}_{\mathrm{KD}}=\sum D_{\mathrm{KL}}\Big(p_{\mathrm{T}}(\cdot\mid x_{\le t})\ \Big\|\ p_{\theta}(\cdot\mid x_{\le t})\Big),
\end{aligned}
\end{equation}
where $p_{\mathrm{T}}$ and $p_\theta$ denote the output logits of the teacher and hybrid student model, and the KL divergence is computed over the full vocabulary.

\textbf{Stage 4: Long-context finetuning.} Following logits distillation, the hybrid model is finetuned on longer sequences using the standard next-token prediction objective:
\begin{equation}
\begin{aligned}
\mathcal{L}_{\mathrm{FT}}=-\sum \log p_{\theta}\big(x_{t+1}\mid x_{\le t}\big).
\end{aligned}
\end{equation}
\subsection{Rethinking FlashMorph}

\textbf{Layer selection objective.} FlashMorph formulates layer selection as a subset selection problem that maximizes hybrid model performance under a fixed full-attention budget~\citep{lan2026morphing}. Since the quality of a selected layers also depends on subsequent recovery training, we evaluate selected architectures based on their post-recovery performance:
\begin{equation}
\begin{aligned}
\label{eq:layer_selection_objective}
\mathcal{I}^\star=\arg\max_{\mathcal{I}\in\mathcal{C}_K}\ \operatorname{Score}\big(\mathcal{M}(\mathcal{I};\theta^\star)\big),
\end{aligned}
\end{equation}
where $\theta^\star$ denotes the model parameters of the selected hybrid architecture $\mathcal{I}$ obtained after initialization from $\theta_0$ and subsequent recovery training, and $\mathrm{Score}(\cdot)$ measures  post-recovery task performance. Individually recovering and evaluating all $\binom{L}{K}$ hybrid architectures is prohibitively expensive. FlashMorph therefore uses Eq.~\ref{eq:flashmorph_stage2_loss} as a proxy for layer selection before recovery. However, this proxy has two \textit{off-policy} limitations that prevent it from fully reflecting the post-recovery performance of the resulting hybrid model:

\textbf{L1: \textit{Off-policy computation}.} During FlashMorph's layer selection, continuous gates mix full- and linear-attention outputs computed from teacher hidden states (Eqs.\ref{eq:flashmorph_mix} and \ref{eq:flashmorph_stage2_loss}). The deployed model instead uses a discrete hybrid architecture under the prescribed full-attention budget, retaining one branch per layer and propagating its own hidden states (Eq.\ref{eq:hybrid_attention}). Thus, rankings learned under continuous mixing can misrepresent layer importance in the discrete hybrid architecture, while teacher-forced layerwise evaluation fails to capture how hidden-state deviations induced by attention replacement propagate through subsequent hybrid layers.

\textbf{L2: \textit{Off-policy weights}.} Eq.~\ref{eq:flashmorph_stage2_loss} performs layer selection by optimizing continuous gates while keeping model weights frozen at their pre-recovery values $\theta_0$, whereas Eq.\ref{eq:layer_selection_objective} evaluates discrete hybrid architectures using the recovered parameters $\theta^\star$. Since the frozen weights have not been adapted to the candidate hybrid architectures, subsequent recovery may alter the layer-importance ranking, potentially rendering the initial selection suboptimal.

These two forms of \textit{off-policy} limitation highlight the need for layer selection to jointly account for end-to-end forward computation and weight adaptation on discrete hybrid architectures satisfying the prescribed full-attention budget. Building on this insight, Sec.~\ref{sec:method}  introduces our proposed FlashMorph2's \textit{on-policy} layer-selection method.
\begin{figure*}[!t]
    \centering
    \includegraphics[width=\linewidth]{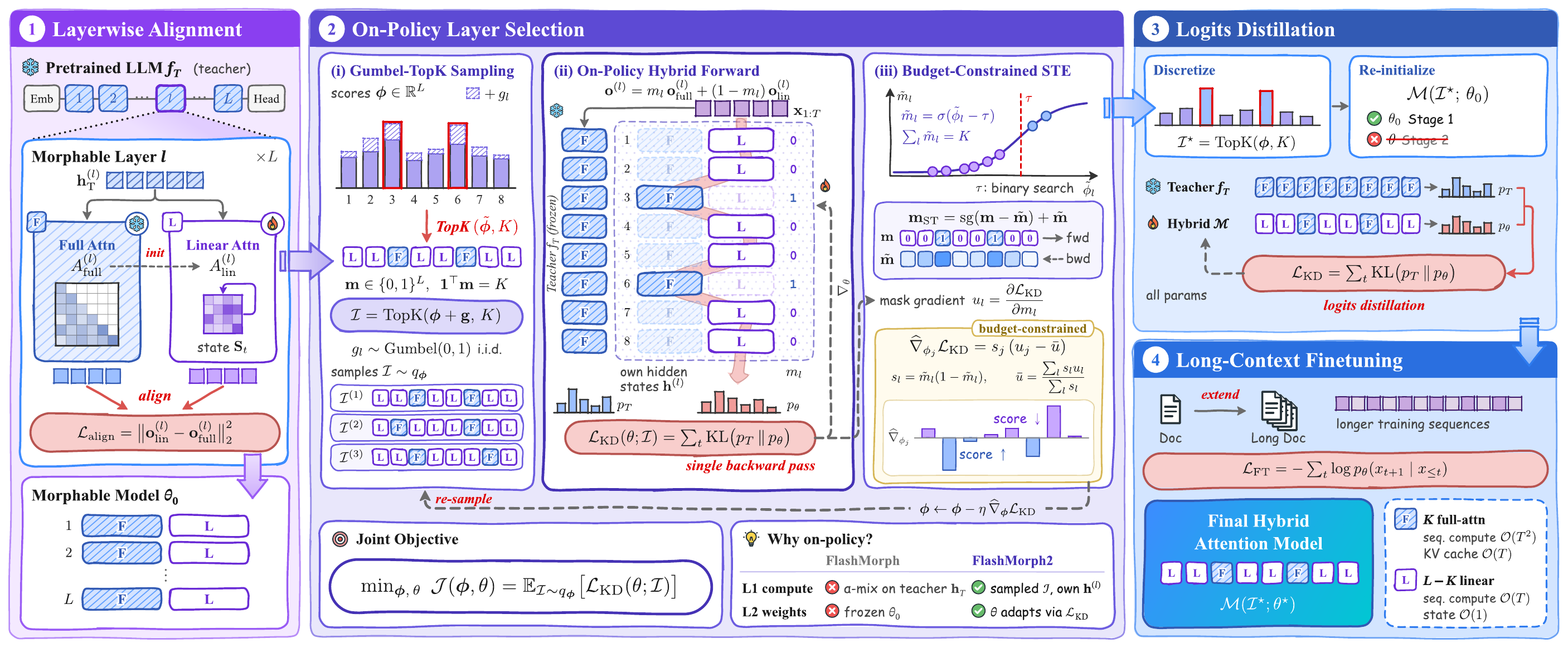}
    \vspace{-1.2em}
    \caption{\textbf{Overview of FlashMorph2.} Following layerwise alignment,  FlashMorph2 constructs a hybrid attention model through \textit{on-policy} layer selection, jointly optimizing layer scores and model weights over sampled discrete hybrid architectures. The resulting hybrid attention model is further trained using logits distillation and long-context finetuning to recover model quality.}
    \label{fig:flashmorph2}
    \vspace{-1.2em}
\end{figure*}

\vspace{-0.3em}
\section{Methodology}
\label{sec:method}
\vspace{-0.2em}
Drawing on \textit{on-policy} learning, FlashMorph2 refines FlashMorph's optimization-based layer selection. Specifically, as illustrated in Fig.~\ref{fig:flashmorph2}, FlashMorph2 samples discrete hybrid configurations satisfying the prescribed full-attention budget, performs end-to-end forward computation with self-propagated hidden states rather than teacher-forced hidden states, and jointly optimizes the layer-selection parameters and model weights using the same training objective as in the subsequent recovery stage to guide the layer selection.

\vspace{-0.3em}
\subsection{On-Policy Hybrid Computation}
\vspace{-0.2em}
We assign each layer a learnable layer-selection score $\phi_l$, collectively forming the vector $\boldsymbol{\phi}\in\mathbb{R}^L$. To explore different hybrid configurations during layer-selection optimization, we add independent standard Gumbel noise to the layer-selection scores and select the $K$ layers with the highest perturbed scores: 
\begin{equation}
\begin{aligned}
\mathcal{I}=\operatorname{TopK}\big(\boldsymbol{\phi}+\mathbf{g},\,K\big),\quad g_l \overset{\text{i.i.d.}}{\sim}\operatorname{Gumbel}(0,1). 
\end{aligned}
\end{equation}
We employ Gumbel-Top-$K$ sampling trick without replacement~\citep{kool2019stochastic}, selecting exactly $K$ distinct full-attention layers per sample, with higher layer-selection scores corresponding to higher selection probabilities. We denote the induced distribution over hybrid architectures in $\mathcal{C}_K$ by $q_{\boldsymbol{\phi}}$, with the corresponding binary mask defined as $m_l=\mathbf{1}[l\in\mathcal{I}]$. All layer-selection scores are initialized to the same constant, ensuring no prior preference for any layer during initial sampling.

Given a sampled mask, the model performs an \textit{on-policy} forward pass through the corresponding discrete hybrid architecture (Eq.~\ref{eq:hybrid_attention}). Each layer operates on hidden states propagated through the sampled architecture and uses the output of its selected attention branch, allowing earlier branch choices to affect subsequent layer inputs and ultimately the final predictions. Layer-selection evaluation thus captures error accumulation and cross-layer compensation induced by attention replacement through forward computation consistent with that of the deployed hybrid architecture.

\subsection{On-Policy Weight Adaptation}

To account for weight adaptation during subsequent recovery, we initialize the student model with the layerwise aligned parameters $\theta_0$ and jointly optimize the layer-selection parameters $\boldsymbol{\phi}$ and model weights $\theta$ over sampled hybrid architectures:
\begin{equation}
\begin{aligned}
\min_{\boldsymbol{\phi},\,\theta}\ \mathcal{J}(\boldsymbol{\phi},\theta)=\mathbb{E}_{\mathcal{I}\sim q_{\boldsymbol{\phi}}}\Big[\mathcal{L}_{\mathrm{KD}}(\theta;\mathcal{I})\Big],
\end{aligned}
\end{equation}
where $\mathcal{L}_{\mathrm{KD}}(\theta;\mathcal{I})$ denotes the logits distillation objective used in subsequent recovery training stage (Eq.~\ref{eq:distill}), computed on the currently sampled hybrid architecture $\mathcal{I}$.
All sampled hybrid architectures are constructed by selecting attention branches from a shared morphable model, directly reusing the parameters of the selected branches. During layer-selection optimization, we compute the distillation loss on sampled architectures and obtain gradients for both the layer-selection parameters $\boldsymbol{\phi}$ and model weights $\theta$ through a single backward pass. The layer-selection parameters are updated via gradient straight-through estimator (STE)~\citep{bengio2013estimating,jang2016categorical} (Sec.~\ref{sec:ste}) to adjust the architecture-sampling distribution. For each sampled architecture, unselected attention branches receive no gradients. As the model weights continuously adapt to the sampled hybrid architectures, the layer-selection parameters are updated based on the corresponding optimization signals. This \textit{on-policy} layer-selection method incorporates weight adaptation into architecture evaluation, mitigating the discrepancy between frozen-weight layer evaluation and post-recovery performance.

After optimization, we remove the sampling noise and select the final full-attention layers as $\mathcal{I}^\star=\operatorname{TopK}(\boldsymbol{\phi},K)$. To isolate the effect of layer selection, we retain only the selected architecture, reinitialize $\mathcal{M}(\mathcal{I}^\star;\theta_0)$ with the original parameters $\theta_0$, and apply the same Stage 3-4 recovery training pipeline. The model weights $\theta$ updated during Stage 2 serve only to guide layer selection and are not used to initialize subsequent recovery training.

\begin{table*}[t]
    \centering
    \caption{\textbf{Long-context Performance on RULER and LongBench.} Avg denotes the average score over 13 RULER or 14 LongBench tasks. For each backbone and linear-attention variant, the \colorbox{cyan!25}{best} and \colorbox{cyan!10}{second-best} scores among layer-selection methods are highlighted.}
    \label{tab:longctx}
    \small
    \setlength{\tabcolsep}{4pt}
    \renewcommand{\arraystretch}{0.88}
    \newcolumntype{Y}{>{\centering\arraybackslash}X}
    \begin{tabularx}{\textwidth}{ll|YYYY|Y|YYYYY|Y}
    \toprule
    \multirow{2}{*}{\textbf{Attn.}} & \multirow{2}{*}{\textbf{Method}} & \multicolumn{5}{c|}{\textbf{RULER}} & \multicolumn{6}{c}{\textbf{LongBench}} \\
    & & NIAH & VT & Agg. & QA & \textbf{Avg} & SQA & MQA & Sum. & FS & Code & \textbf{Avg} \\
    \midrule

    \multicolumn{13}{l}{\textit{0.6B Backbone}} \\
    \rowcolor{gray!15}
    Full & Qwen3 & 94.38 & 95.32 & 55.78 & 33.56 & 79.15 & 21.22 & 20.86 & 19.39 & 60.78 & 33.26 & 30.95 \\

    \midrule
    \multirow{5}{*}{Lightning}
    & Uniform & 61.12 & 1.56 & 51.64 & 29.99 & 50.29 & 19.90 & 16.46 & 17.92 & \cellcolor{cyan!25}48.58 & 33.28 & 26.80 \\
    & KL-LS & 54.49 & 2.20 & 34.46 & 31.66 & 43.87 & 18.52 & \cellcolor{cyan!25}20.15 & 18.60 & 43.32 & \cellcolor{cyan!10}38.71 & 27.09 \\
    & HALO & 59.51 & 1.20 & 50.48 & 32.61 & 49.50 & 19.05 & 18.24 & 17.11 & 44.72 & 34.65 & 26.19 \\
    & FlashMorph & \cellcolor{cyan!25}68.56 & \cellcolor{cyan!10}10.84 & \cellcolor{cyan!10}55.03 & \cellcolor{cyan!10}35.33 & \cellcolor{cyan!10}56.93 & \cellcolor{cyan!25}23.36 & 19.16 & \cellcolor{cyan!10}19.55 & 46.43 & 36.73 & \cellcolor{cyan!10}28.50 \\
    & \ourmethod & \cellcolor{cyan!10}68.52 & \cellcolor{cyan!25}17.24 & \cellcolor{cyan!25}58.46 & \cellcolor{cyan!25}36.49 & \cellcolor{cyan!25}58.10 & \cellcolor{cyan!10}22.90 & \cellcolor{cyan!10}19.39 & \cellcolor{cyan!25}19.80 & \cellcolor{cyan!10}47.34 & \cellcolor{cyan!25}39.97 & \cellcolor{cyan!25}29.16 \\

    \midrule
    \multirow{5}{*}{GLA}
    & Uniform & 62.16 & 4.44 & 48.80 & 27.13 & 50.27 & 19.77 & 15.61 & 16.82 & \cellcolor{cyan!10}46.88 & 29.56 & 25.45 \\
    & KL-LS & 67.95 & 1.80 & 43.47 & 29.95 & 53.25 & 21.86 & 17.64 & \cellcolor{cyan!10}19.19 & \cellcolor{cyan!25}47.91 & \cellcolor{cyan!10}30.79 & \cellcolor{cyan!10}27.24 \\
    & HALO & 65.84 & 2.08 & 47.37 & 29.98 & 52.58 & 20.03 & \cellcolor{cyan!25}19.23 & 17.37 & 46.83 & 29.00 & 26.31 \\
    & FlashMorph & \cellcolor{cyan!10}68.99 & \cellcolor{cyan!10}5.52 & \cellcolor{cyan!10}51.27 & \cellcolor{cyan!10}32.56 & \cellcolor{cyan!10}55.78 & \cellcolor{cyan!25}23.64 & \cellcolor{cyan!10}18.99 & 18.40 & 43.72 & 29.64 & 26.68 \\
    & \ourmethod & \cellcolor{cyan!25}76.16 & \cellcolor{cyan!25}7.20 & \cellcolor{cyan!25}52.62 & \cellcolor{cyan!25}36.62 & \cellcolor{cyan!25}61.15 & \cellcolor{cyan!10}23.49 & 17.93 & \cellcolor{cyan!25}19.46 & 46.42 & \cellcolor{cyan!25}31.98 & \cellcolor{cyan!25}27.57 \\

    \midrule
    \multirow{5}{*}{GDN}
    & Uniform & 64.09 & 3.76 & 50.23 & 33.26 & 52.57 & 21.19 & 16.40 & 17.89 & 49.09 & 35.19 & 27.43 \\
    & KL-LS & 71.10 & 2.96 & 48.97 & 34.76 & 56.86 & 20.89 & 17.25 & \cellcolor{cyan!25}20.35 & \cellcolor{cyan!25}49.94 & \cellcolor{cyan!25}38.85 & 28.79 \\
    & HALO & 60.20 & \cellcolor{cyan!25}20.28 & 44.81 & 33.44 & 50.64 & 20.78 & 16.24 & 18.48 & 46.47 & 36.97 & 27.13 \\
    & FlashMorph & \cellcolor{cyan!10}72.48 & 15.60 & \cellcolor{cyan!10}56.54 & \cellcolor{cyan!25}37.65 & \cellcolor{cyan!10}60.29 & \cellcolor{cyan!10}23.37 & \cellcolor{cyan!10}18.91 & 19.68 & 48.78 & \cellcolor{cyan!10}37.94 & \cellcolor{cyan!10}29.15 \\
    & \ourmethod & \cellcolor{cyan!25}76.83 & \cellcolor{cyan!10}18.68 & \cellcolor{cyan!25}60.54 & \cellcolor{cyan!10}36.49 & \cellcolor{cyan!25}63.65 & \cellcolor{cyan!25}24.21 & \cellcolor{cyan!25}20.04 & \cellcolor{cyan!10}20.15 & \cellcolor{cyan!10}49.29 & 37.88 & \cellcolor{cyan!25}29.77 \\

    \midrule
    \midrule

    \multicolumn{13}{l}{\textit{1.7B Backbone}} \\
    \rowcolor{gray!15}
    Full & Qwen3 & 99.51 & 95.72 & 77.47 & 46.44 & 87.66 & 26.23 & 13.12 & 18.53 & 66.02 & 41.14 & 32.43 \\

    \midrule
    \multirow{6}{*}{Lightning}
    & Uniform & 71.36 & 29.16 & 51.97 & 40.73 & 60.41 & \cellcolor{cyan!10}21.19 & 18.24 & 20.27 & 57.69 & 37.59 & 30.53 \\
    & PostNAS & 68.17 & 23.16 & \cellcolor{cyan!25}64.94 & 42.88 & 60.32 & 20.69 & \cellcolor{cyan!10}22.06 & 19.92 & 55.39 & 36.81 & 30.56 \\
    & KL-LS & 56.95 & 0.24 & 44.01 & 37.22 & 47.56 & 20.07 & 18.09 & \cellcolor{cyan!10}21.11 & 56.05 & \cellcolor{cyan!10}42.42 & 30.77 \\
    & HALO & 72.89 & 29.48 & \cellcolor{cyan!10}63.60 & 43.91 & 63.66 & 20.31 & 21.42 & 20.30 & 57.92 & 40.12 & \cellcolor{cyan!10}31.44 \\
    & FlashMorph & \cellcolor{cyan!10}77.59 & \cellcolor{cyan!10}31.64 & 60.05 & \cellcolor{cyan!10}45.01 & \cellcolor{cyan!10}66.35 & 20.25 & 21.06 & 20.02 & \cellcolor{cyan!10}59.23 & 37.09 & 31.13 \\
    & \ourmethod & \cellcolor{cyan!25}79.96 & \cellcolor{cyan!25}41.24 & 62.91 & \cellcolor{cyan!25}47.74 & \cellcolor{cyan!25}69.40 & \cellcolor{cyan!25}22.71 & \cellcolor{cyan!25}23.56 & \cellcolor{cyan!25}21.30 & \cellcolor{cyan!25}59.72 & \cellcolor{cyan!25}42.53 & \cellcolor{cyan!25}33.35 \\

    \midrule
    \multirow{6}{*}{GLA}
    & Uniform & 72.86 & 20.56 & 54.98 & 41.98 & 61.34 & 21.00 & 18.53 & \cellcolor{cyan!25}18.73 & 57.56 & 38.49 & 30.32 \\
    & PostNAS & 74.95 & 19.76 & 51.59 & \cellcolor{cyan!10}45.92 & 62.64 & \cellcolor{cyan!25}24.23 & \cellcolor{cyan!25}21.76 & 17.21 & 56.74 & 38.77 & \cellcolor{cyan!10}31.24 \\
    & KL-LS & 58.94 & 11.60 & 50.48 & 38.59 & 50.86 & 19.91 & 20.30 & \cellcolor{cyan!10}17.80 & 57.45 & 43.69 & 30.98 \\
    & HALO & 64.45 & \cellcolor{cyan!10}25.84 & \cellcolor{cyan!25}62.20 & 37.39 & 56.97 & 22.38 & 17.85 & 15.93 & 56.19 & 40.59 & 29.87 \\
    & FlashMorph & \cellcolor{cyan!25}80.43 & \cellcolor{cyan!25}30.52 & 56.48 & 42.83 & \cellcolor{cyan!10}67.12 & 21.93 & 19.16 & 15.77 & \cellcolor{cyan!10}58.89 & \cellcolor{cyan!25}43.82 & 31.06 \\
    & \ourmethod & \cellcolor{cyan!10}79.14 & 22.28 & \cellcolor{cyan!10}61.81 & \cellcolor{cyan!25}48.01 & \cellcolor{cyan!25}67.31 & \cellcolor{cyan!10}23.38 & \cellcolor{cyan!10}20.50 & 15.84 & \cellcolor{cyan!25}60.20 & \cellcolor{cyan!10}43.79 & \cellcolor{cyan!25}31.95 \\

    \midrule
    \multirow{6}{*}{GDN}
    & Uniform & 75.56 & 63.32 & 59.41 & 45.42 & 67.50 & 19.53 & \cellcolor{cyan!10}22.27 & 20.48 & 59.23 & 35.41 & 31.10 \\
    & PostNAS & 79.20 & 53.16 & 64.97 & 44.92 & 69.73 & \cellcolor{cyan!10}25.24 & 20.72 & \cellcolor{cyan!25}21.73 & 58.16 & 36.65 & 32.20 \\
    & KL-LS & 79.71 & \cellcolor{cyan!25}68.12 & 65.44 & 47.91 & 71.73 & 24.57 & 19.74 & \cellcolor{cyan!10}21.41 & \cellcolor{cyan!10}59.62 & \cellcolor{cyan!10}41.06 & \cellcolor{cyan!10}32.73 \\
    & HALO & 76.18 & 62.68 & 63.31 & 46.94 & 68.66 & 25.13 & \cellcolor{cyan!25}22.30 & 18.55 & 57.46 & 40.94 & 32.30 \\
    & FlashMorph & \cellcolor{cyan!10}81.79 & \cellcolor{cyan!10}66.76 & \cellcolor{cyan!10}65.89 & \cellcolor{cyan!10}48.52 & \cellcolor{cyan!10}73.07 & 24.69 & 22.23 & 20.78 & 59.51 & 37.52 & 32.62 \\
    & \ourmethod & \cellcolor{cyan!25}84.83 & 58.80 & \cellcolor{cyan!25}66.28 & \cellcolor{cyan!25}48.96 & \cellcolor{cyan!25}74.46 & \cellcolor{cyan!25}25.72 & 20.53 & 21.39 & \cellcolor{cyan!25}61.16 & \cellcolor{cyan!25}43.48 & \cellcolor{cyan!25}33.81 \\

    \bottomrule
    \end{tabularx}
    \vspace{-1.0em}
\end{table*}

\subsection{Budget-Constrained Straight-Through Estimator}
\label{sec:ste}

Discrete Top-$K$ selection is non-differentiable. To estimate gradients for the layer-selection parameters, we construct a continuous relaxation that preserves the same budget constraint~\citep{wijk2026differentiable}. For a sampled hybrid architecture, let $\tilde{\boldsymbol{\phi}}=\boldsymbol{\phi}+\mathbf{g}$ denote the perturbed layer-selection scores, and define:
\begin{equation}
\begin{aligned}
\tilde m_l=\sigma\big(\tilde\phi_l-\tau\big),\quad
\sum_{l=1}^{L}\sigma\big(\tilde\phi_l-\tau\big)=K,
\end{aligned}
\end{equation}
where $\sigma$ denotes the sigmoid function, and the threshold $\tau$ is determined via binary search. We use the binary mask $\mathbf{m}$ in the forward pass during layer selection stage and propagate gradients through the continuous relaxation $\tilde{\mathbf{m}}$ in the backward pass, following the gradient straight-through estimator (STE)~\citep{bengio2013estimating,jang2016categorical}:
\begin{equation}
\begin{aligned}
\mathbf{m}_{\mathrm{ST}}=\operatorname{sg}\big(\mathbf{m}-\tilde{\mathbf{m}}\big)+\tilde{\mathbf{m}},
\end{aligned}
\end{equation}
where $\operatorname{sg}(\cdot)$ denotes the stop-gradient operator. Although the threshold is computed outside the automatic differentiation graph, its dependence on the selection scores is accounted for through implicit differentiation during backpropagation. Let $u_l=\partial\mathcal{L}_{\mathrm{KD}}/\partial m_l$ denote the gradient of the loss with respect to the mask, and define $s_l=\tilde m_l(1-\tilde m_l)$. We obtain the following biased gradient straight-through estimator for the expected discrete objective:
\begin{equation}
\begin{aligned}
\widehat{\nabla}_{\phi_j}\mathcal{L}_{\mathrm{KD}}=s_j\big(u_j-\bar u\big),\quad
\bar u=\frac{\sum_{l=1}^{L}s_lu_l}{\sum_{l=1}^{L}s_l}.
\end{aligned}
\end{equation}
Implicit differentiation of the threshold $\tau$ couples gradients with respect to layer-selection scores under the full-attention budget constraint. Computing $u_l$ requires the gradient backpropagated from the end-to-end loss and the difference between full- and linear-attention outputs. Both branches are therefore evaluated, while the binary mask ensures that only the selected branch contributes to the forward output.


\begin{table*}[t]
    \centering
    \caption{\textbf{Performance on Recall-intensive Tasks.} Avg denotes the average score across all six recall-intensive tasks. For each backbone and linear-attention variant, the \colorbox{cyan!25}{best} and \colorbox{cyan!10}{second-best} scores among layer-selection methods are highlighted.}
    \label{tab:recall}
    \small
    \setlength{\tabcolsep}{1.25pt}
    \renewcommand{\arraystretch}{1.09}
    \newcolumntype{Z}{>{\centering\arraybackslash}X}
    \begin{tabularx}{\textwidth}{ll|*{6}{Z}|Z|*{6}{Z}|Z}
    \toprule
    \multirow{2}{*}{\textbf{Attn.}} & \multirow{2}{*}{\textbf{Method}}
    & \multicolumn{7}{c|}{\textbf{\textit{0.6B Backbone}}} & \multicolumn{7}{c}{\textbf{\textit{1.7B Backbone}}} \\
    \cmidrule(lr){3-9} \cmidrule(lr){10-16}
    & & FDA & SWDE & SQD & NQ & TQA & DROP & \textbf{Avg}
    & FDA & SWDE & SQD & NQ & TQA & DROP & \textbf{Avg} \\
    \midrule
    \rowcolor{gray!15}
    Full & Qwen3 & 76.95 & 66.52 & 46.35 & 30.92 & 57.41 & 28.27 & 51.07 & 70.78 & 71.11 & 51.21 & 34.75 & 63.39 & 30.81 & 53.67 \\

    \midrule
    \multirow{6}{*}{Lightning} & Uniform & 54.90 & \cellcolor{cyan!25}61.93 & 45.58 & \cellcolor{cyan!10}25.34 & 57.76 & 27.17 & 45.45 & 49.36 & 62.83 & 50.97 & 27.91 & \cellcolor{cyan!10}64.34 & 31.53 & 47.82 \\
     & PostNAS & \textemdash & \textemdash & \textemdash & \textemdash & \textemdash & \textemdash & \textemdash & 54.08 & \cellcolor{cyan!25}64.81 & 51.58 & 28.51 & 64.10 & 32.34 & 49.24 \\
     & KL-LS & 49.36 & 57.61 & 46.92 & 22.68 & \cellcolor{cyan!10}58.35 & 27.93 & 43.81 & 48.64 & 60.13 & 50.94 & \cellcolor{cyan!10}29.17 & 63.15 & 32.20 & 47.37 \\
     & HALO & \cellcolor{cyan!10}56.72 & 61.21 & \cellcolor{cyan!10}46.95 & 23.25 & 58.12 & 28.17 & \cellcolor{cyan!10}45.74 & 49.36 & \cellcolor{cyan!10}64.27 & 51.01 & 28.13 & 63.57 & \cellcolor{cyan!10}33.01 & 48.23 \\
     & FlashMorph & 53.27 & \cellcolor{cyan!10}61.57 & 46.65 & 24.87 & \cellcolor{cyan!25}59.12 & \cellcolor{cyan!25}28.89 & 45.73 & \cellcolor{cyan!10}57.17 & 62.83 & \cellcolor{cyan!10}51.91 & 27.91 & \cellcolor{cyan!10}64.34 & \cellcolor{cyan!25}33.72 & \cellcolor{cyan!10}49.65 \\
     & \ourmethod & \cellcolor{cyan!25}59.35 & \cellcolor{cyan!10}61.57 & \cellcolor{cyan!25}47.92 & \cellcolor{cyan!25}25.94 & 56.87 & \cellcolor{cyan!10}28.80 & \cellcolor{cyan!25}46.74 & \cellcolor{cyan!25}60.16 & \cellcolor{cyan!10}64.27 & \cellcolor{cyan!25}52.65 & \cellcolor{cyan!25}30.00 & \cellcolor{cyan!25}64.45 & 32.06 & \cellcolor{cyan!25}50.60 \\

    \midrule
    \multirow{6}{*}{GLA} & Uniform & 54.08 & 61.93 & 46.25 & 25.37 & 57.94 & \cellcolor{cyan!10}28.94 & 45.75 & 53.54 & 63.01 & 51.51 & 29.17 & 65.23 & 32.82 & 49.21 \\
     & PostNAS & \textemdash & \textemdash & \textemdash & \textemdash & \textemdash & \textemdash & \textemdash & 57.62 & \cellcolor{cyan!25}66.07 & 51.98 & 29.90 & 65.17 & 32.82 & 50.59 \\
     & KL-LS & 57.62 & 60.94 & \cellcolor{cyan!10}48.56 & \cellcolor{cyan!10}26.35 & \cellcolor{cyan!10}58.89 & 28.17 & 46.76 & 51.91 & 64.18 & 51.44 & 29.49 & 64.93 & 30.86 & 48.80 \\
     & HALO & 60.25 & 60.31 & 47.65 & 24.77 & 58.00 & \cellcolor{cyan!25}29.80 & 46.80 & 49.73 & 59.50 & 50.80 & 27.75 & 64.75 & 31.62 & 47.36 \\
     & FlashMorph & \cellcolor{cyan!10}62.98 & \cellcolor{cyan!10}62.20 & 46.75 & 26.13 & 57.88 & 27.93 & \cellcolor{cyan!10}47.31 & \cellcolor{cyan!25}61.98 & 61.75 & \cellcolor{cyan!25}53.35 & \cellcolor{cyan!25}30.66 & \cellcolor{cyan!10}65.46 & \cellcolor{cyan!25}33.92 & \cellcolor{cyan!10}51.19 \\
     & \ourmethod & \cellcolor{cyan!25}64.16 & \cellcolor{cyan!25}65.35 & \cellcolor{cyan!25}48.76 & \cellcolor{cyan!25}27.37 & \cellcolor{cyan!25}59.24 & 28.13 & \cellcolor{cyan!25}48.84 & \cellcolor{cyan!10}59.26 & \cellcolor{cyan!10}65.44 & \cellcolor{cyan!10}53.25 & \cellcolor{cyan!10}29.97 & \cellcolor{cyan!25}66.11 & \cellcolor{cyan!10}33.25 & \cellcolor{cyan!25}51.21 \\

    \midrule
    \multirow{6}{*}{GDN} & Uniform & 55.08 & 62.38 & 45.81 & 25.94 & 57.88 & 28.80 & 45.98 & 52.72 & 62.38 & 52.38 & 29.36 & 65.76 & 32.10 & 49.12 \\
     & PostNAS & \textemdash & \textemdash & \textemdash & \textemdash & \textemdash & \textemdash & \textemdash & 59.17 & \cellcolor{cyan!25}67.51 & 53.18 & \cellcolor{cyan!25}32.28 & 65.46 & 33.78 & \cellcolor{cyan!10}51.90 \\
     & KL-LS & \cellcolor{cyan!25}64.25 & 62.29 & 47.92 & 26.42 & 58.77 & 27.55 & 47.87 & \cellcolor{cyan!25}61.89 & 64.36 & 53.08 & \cellcolor{cyan!10}32.15 & 65.28 & 32.06 & 51.47 \\
     & HALO & 57.17 & 62.29 & \cellcolor{cyan!10}48.09 & 26.13 & \cellcolor{cyan!25}59.60 & \cellcolor{cyan!10}29.28 & 47.09 & 60.89 & 64.90 & 52.71 & 31.17 & \cellcolor{cyan!10}66.29 & \cellcolor{cyan!10}33.88 & 51.64 \\
     & FlashMorph & 62.34 & \cellcolor{cyan!10}63.73 & \cellcolor{cyan!25}48.63 & \cellcolor{cyan!10}27.24 & \cellcolor{cyan!10}59.00 & 27.17 & \cellcolor{cyan!10}48.02 & 59.53 & 65.17 & \cellcolor{cyan!10}53.25 & 32.06 & 66.00 & \cellcolor{cyan!25}34.64 & 51.77 \\
     & \ourmethod & \cellcolor{cyan!10}63.43 & \cellcolor{cyan!25}64.18 & 47.39 & \cellcolor{cyan!25}27.78 & 58.77 & \cellcolor{cyan!25}29.47 & \cellcolor{cyan!25}48.50 & \cellcolor{cyan!10}61.71 & \cellcolor{cyan!10}66.88 & \cellcolor{cyan!25}53.62 & 31.58 & \cellcolor{cyan!25}66.94 & 33.21 & \cellcolor{cyan!25}52.32 \\

    \bottomrule
    \end{tabularx}
\end{table*}

\section{Experiments}

\subsection{Experiment Setup}
\textbf{Model and training.} We use Qwen3-0.6B and Qwen3-1.7B~\citep{yang2025qwen3} as pretrained Transformer backbones and construct hybrid models with three linear-attention variants: Lightning Attention~\citep{qin2024lightning}, Gated Linear Attention (GLA)~\citep{yang2023gated}, and Gated DeltaNet (GDN)~\citep{yang2025gated}. We implement the linear-attention branches using the \texttt{flash-\allowbreak linear-\allowbreak attention} library~\citep{yang2024fla}, incorporating QK normalization, GQA-to-MHA conversion, output normalization, and an output gate. Both full- and linear-attention layers adopt RoPE~\citep{su2024roformer}. Unless otherwise specified, the hybrid ratio of linear-attention to full-attention layers is 3:1. We follow the four-stage Transformer-to-hybrid conversion pipeline described in Sec.~\ref{sec:pipeline}, using training data randomly sampled from DCLM~\citep{li2024datacomp}. Details on model and training configurations are in Appendix~\ref{app:conf}.

\textbf{Evaluation.} We evaluate FlashMorph2 on long-context capabilities using RULER~\citep{hsieh2024ruler} and 14 tasks from LongBench~\citep{bai2024longbench}. We further evaluate performance on 6 recall-intensive tasks using the \texttt{prefix-\allowbreak linear-\allowbreak attention} evaluation suite~\citep{arora2024just}. We also evaluate language modeling perplexity and zero-shot commonsense reasoning using \texttt{lm-\allowbreak evaluation-\allowbreak harness}~\citep{eval-harness}. All evaluations are conducted using vLLM~\citep{kwon2023efficient} for accelerated inference. Detailed evaluation settings are also provided in the Appendix~\ref{app:eval}.

\textbf{Baselines.} We compare FlashMorph2 against five layer selection  baselines, including: uniform interleaving (Uniform), PostNAS~\citep{gu2026jet}, KL-LS~\citep{li2026distilling}, HALO~\citep{chen2026hybrid}, and FlashMorph~\citep{lan2026morphing}. To isolate the effect of layer selection, all methods use the same full-attention budget, layerwise-aligned initialization, and subsequent recovery training settings.

\subsection{Main Results}

\textbf{Long-context capabilities.} Table~\ref{tab:longctx} reports long-context performance under the same full-attention budget. FlashMorph2 achieves the highest average scores on both RULER and LongBench across all six combinations of backbone size and linear-attention variant. Averaged across these configurations, it raises the RULER score from 63.26 to 65.68 ($+2.42$ points) and the LongBench score from 29.86 to 30.94 ($+1.08$ points) compared with FlashMorph. These results demonstrate that FlashMorph2 identifies hybrid architectures with consistently superior long-context performance following an identical recovery training settings.

\textbf{Recall-intensive tasks.} Table~\ref{tab:recall} reports performance on six recall-intensive tasks. FlashMorph2 achieves the highest average score across tasks in all six combinations of backbone size and linear-attention variant. Averaged across these configurations, it improves the score from 48.94 to 49.70 ($+0.76$ points) compared with FlashMorph. These results indicate that the benefits of FlashMorph2 extend to real-world in-context retrieval tasks.

\textbf{Language modeling and commonsense reasoning.} Table~\ref{tab:zero_shot} in the Appendix reports language modeling perplexity and zero-shot commonsense reasoning accuracy. FlashMorph2 remains competitive with the layer-selection baselines on these evaluations, suggesting that general language capabilities are preserved alongside the improvements in long-context and recall-intensive tasks.

\begin{figure*}[t]
    \centering
    \begin{minipage}[t]{0.492\textwidth}
        \centering
        \includegraphics[width=\linewidth]{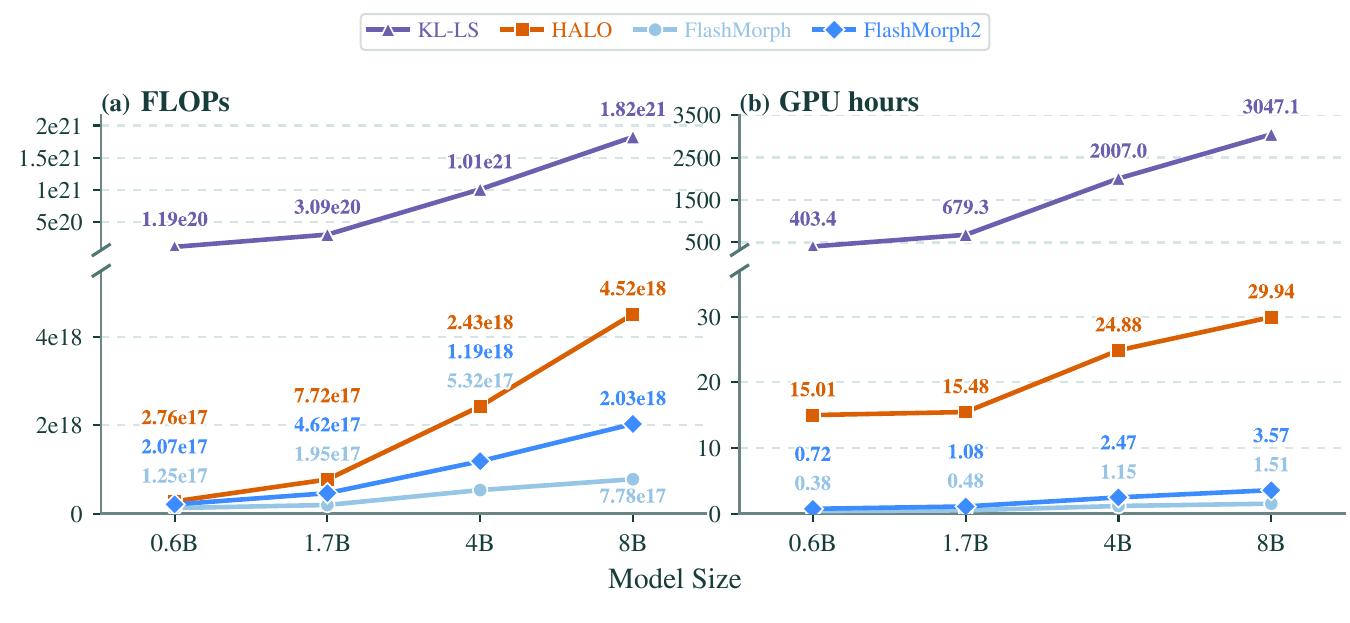}
        \vspace{-1.8em}
        \caption{\textbf{Layer-selection Efficiency across Model Scales.} Comparison of (a) FLOPs and (b) GPU hours required for layer selection across Qwen3 models from 0.6B to 8B parameters.}
        \label{fig:efficiency}
    \end{minipage}
    \hfill
    \begin{minipage}[t]{0.492\textwidth}
        \centering
        \includegraphics[width=\linewidth]{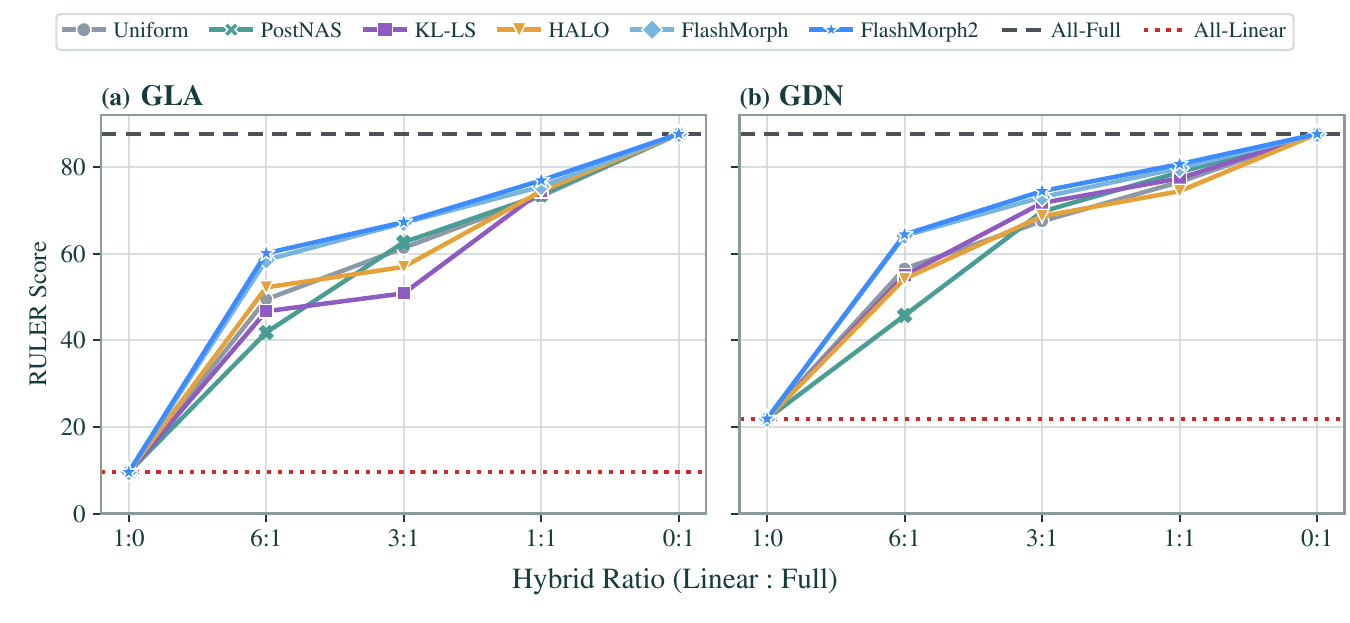}
        \vspace{-1.8em}
        \caption{\textbf{RULER Performance across Hybrid Ratios.} Average RULER scores on the Qwen3-1.7B backbone across varying hybrid ratios for (a) GLA and (b) GDN variants.}
        \label{fig:ruler_hybrid_ratio}
    \end{minipage}
\end{figure*}

\subsection{Analysis}

\textbf{Layer-selection efficiency.}
Fig.~\ref{fig:efficiency} reports the computational cost of layer selection across different model scales. Compared with FlashMorph, FlashMorph2 requires $1.7$--$2.6\times$ the FLOPs and $1.9$--$2.4\times$ the GPU hours, reflecting the additional computation required for end-to-end evaluation of discrete hybrid architectures and concurrent model adaptation. This overhead is confined to a one-time selection stage, which accounts for approximately only $<1\%$ of the overall conversion token budget, and does not affect deployment efficiency under the same full-attention budget. Across models ranging from 0.6B to 8B parameters, FlashMorph2 completes layer selection in only $0.72$--$3.57$ GPU hours. It is also substantially more efficient than layerwise-scoring approaches. Relative to FlashMorph2, HALO requires $1.3$--$2.2\times$ as many FLOPs and $8.4$--$20.8\times$ as many GPU hours, while KL-LS requires $575$--$897\times$ and $560$--$854\times$ as many, respectively. Overall, despite its modest overhead over FlashMorph, FlashMorph2 remains highly efficient and scalable while being substantially more efficient than alternative layer-selection methods.

\textbf{Hybrid ratios.} Fig.~\ref{fig:ruler_hybrid_ratio} reports RULER performance on Qwen3-1.7B across different hybrid ratios for GLA and GDN. While full linearization ($1{:}0$) reduces the RULER score from 87.66 to 9.57 on GLA and 21.89 on GDN, retaining a small subset of full-attention layers substantially restores retrieval performance. Across all evaluated hybrid ratios ($6{:}1$, $3{:}1$, and $1{:}1$), FlashMorph2 consistently outperforms all baselines on both linear-attention variants. The advantage of optimization-based layer selection is particularly pronounced under the tightest full-attention budget ($6{:}1$), where FlashMorph2 achieves 60.12 on GLA and 64.45 on GDN, outperforming Uniform by $+10.65$ and $+7.86$ points while several competing methods fall below simple uniform interleaving. As the full-attention ratio increases to $3{:}1$ and $1{:}1$, performance across methods gradually converges toward the full-attention backbone, yet FlashMorph2 continues to improve upon FlashMorph in all settings, reaching 76.93 on GLA ($+1.42$) and 80.69 on GDN ($+0.92$) at $1{:}1$. Overall, these results confirm that FlashMorph2 remains effective and robust across varying hybrid ratios.

\textbf{Ablation study.} Table~\ref{tab:ablation} evaluates the effects of \textit{on-policy} computation and \textit{on-policy} weight adaptation on Qwen3-1.7B with GDN. Removing \textit{on-policy} computation significantly reduces scores by $29.55$ points on RULER and $5.71$ points on LongBench. Inspecting the selected architectures reveals that this variant allocates all seven full-attention layers to the deepest layers $\{20, 21, 23, 24, 25, 26, 27\}$, leaving the first 20 layers entirely linear. This concentration reflects the mismatch between teacher-forced local scoring and end-to-end discrete hybrid computation, where early hidden-state deviations accumulate across layers. Removing \textit{on-policy} weight adaptation instead lowers RULER by $4.88$ points and LongBench by $1.38$ points, replacing layers $\{6, 17, 19, 20\}$ selected by FlashMorph2 with $\{0, 1, 2, 26\}$, which indicates that weight adaptation substantially alters the preferred budget allocation. Removing both components reverts to FlashMorph, which still trails FlashMorph2 by $1.39$ points on RULER and $1.19$ points on LongBench. Overall, these results confirm that discrete hybrid computation and weight adaptation are complementary and must be jointly considered during layer selection.

\begin{table}[!t]
    \centering
    \caption{\textbf{Ablation Study of FlashMorph2.} Average RULER and LongBench scores on the Qwen3-1.7B backbone with GDN.}
    \label{tab:ablation}
    \small
    \setlength{\tabcolsep}{3.5pt}
    \renewcommand{\arraystretch}{1.0}
    \begin{tabularx}{0.92\columnwidth}{l|*{2}{>{\centering\arraybackslash}X}}
    \toprule
    \textbf{Method}
    & \textbf{RULER}
    & \textbf{LongBench} \\
    \midrule
    FlashMorph2 & 74.46 & 33.81 \\
    \midrule
    \textit{w/o} On-Policy Computation & 44.91 & 28.10 \\
    \textit{w/o} On-Policy Weight & 69.58 & 32.43 \\
    \textit{w/o} Both (FlashMorph) & 73.07 & 32.62 \\
    \bottomrule
    \end{tabularx}
    \vspace{-1.4em}
\end{table}
\section{Conclusion}

In this work, we present FlashMorph2, an \textit{on-policy} layer-selection method for converting pretrained Transformers into hybrid attention models under a prescribed full-attention budget. To overcome the \textit{off-policy} computation and weight mismatches in prior optimization-based selection, FlashMorph2 samples discrete budget-constrained hybrid architectures, performs end-to-end forward passes with self-propagated hidden states, and jointly optimizes layer-selection scores and model weights under the recovery training objective. Experiments across multiple Qwen3 model scales, linear-attention variants, and hybrid ratios show that FlashMorph2 consistently improves long-context and recall-intensive performance over existing baselines while maintaining high selection efficiency and scalability.

\bibliography{example_paper}
\bibliographystyle{icml2027}

\appendix
\onecolumn

\section{Experimental Details}
\label{app:conf}

\subsection{Model Configurations}

Table~\ref{tab:model_config} summarizes the model architectural specifications of the Qwen3 backbones~\citep{yang2025qwen3} and their corresponding hybrid attention models under the default $3{:}1$ linear-to-full attention hybrid ratio. To convert grouped-query attention (GQA) in the pretrained full-attention layers into multi-head linear attention (MHA), we expand the pretrained key and value projection weights by repeating each KV head across its corresponding query group during initialization, while directly reusing the pretrained query and output projection weights. Each linear-attention branch is further equipped with RMSNorm on queries, keys, and branch outputs, an element-wise sigmoid output gate initialized to zero, and rotary position embeddings (RoPE)~\citep{su2024roformer}. All newly introduced parameters in the linear-attention branches (e.g., decay/forget gates and output gates) are trained from scratch starting in Stage 1.

Table~\ref{tab:layer_selection_results} presents the complete layer-importance rankings and selected full-attention layers produced by FlashMorph2 and baseline layer-selection methods across different backbones and linear-attention variants. Layer indices are zero-based, with boxed indices indicating the layers that retain full attention and all remaining layers replaced by linear attention.

\begin{table*}[!htbp]
    \centering
    \caption{\textbf{Hybrid Attention Model Configurations across Different Scales.}}
    \label{tab:model_config}
    \small
    \begin{tabularx}{0.62\textwidth}{l|*{4}{>{\centering\arraybackslash}X}}
    \toprule
    \textbf{Attribute} & \textbf{0.6B} & \textbf{1.7B} & \textbf{8B} & \textbf{30B-A3B} \\
    \midrule
    \#layers & 28 & 28 & 36 & 48 \\
    \#full-attention layers & 7 & 7 & 9 & 12 \\
    \#linear-attention layers & 21 & 21 & 27 & 36 \\
    hidden dimension & 1024 & 2048 & 4096 & 2048 \\
    dense FFN dimension & 3072 & 6144 & 12288 & \textemdash \\
    \#experts & \textemdash & \textemdash & \textemdash & 128 \\
    \#active experts & \textemdash & \textemdash & \textemdash & 8 \\
    expert FFN dimension & \textemdash & \textemdash & \textemdash & 768 \\
    head dimension & 128 & 128 & 128 & 128 \\
    \#query heads & 16 & 16 & 32 & 32 \\
    \#full-attention KV heads & 8 & 8 & 8 & 4 \\
    \#linear-attention KV heads & 16 & 16 & 32 & 32 \\
    \bottomrule
\end{tabularx}
\end{table*}

\subsection{Training Configurations}

Table~\ref{tab:training_config} summarizes the training hyperparameters across all four stages of the Transformer-to-hybrid conversion pipeline: layerwise alignment (Stage 1), \textit{on-policy} layer selection (Stage 2), logits distillation (Stage 3), and long-context finetuning (Stage 4). All stages are trained on randomly sampled subsets of DCLM~\citep{li2024datacomp} using the Adam optimizer~\citep{kingma2014adam} without weight decay. In Stage 1, only the linear-attention branches are updated while all pretrained backbone weights remain frozen. In Stage 2, we initialize all layer-selection scores $\boldsymbol{\phi}$ to $0.5$ and jointly optimize $\boldsymbol{\phi}$ and the model weights $\theta$ under a Warmup--Stable--Decay (WSD)~\citep{hu2024minicpm} learning rate schedule. After Stage 2, we retain the top-$K$ full-attention layers, reinitialize the selected hybrid attention model with the Stage 1 weights $\theta_0$, and train all model parameters end to end during Stage 3 and Stage 4. To ensure a fair comparison, all layer-selection methods share the same Stage 1 checkpoint and identical Stage 3--4 recovery training settings.

\begin{table*}[!htbp]
    \centering
    \caption{\textbf{Training Configurations across the Four-Stage Transformer-to-hybrid Conversion Pipeline.}}
    \label{tab:training_config}
    \small
    \begin{tabularx}{0.77\textwidth}{l|*{4}{>{\centering\arraybackslash}X}}
    \toprule
    \textbf{Hyperparameter} & \textbf{Stage 1} & \textbf{Stage 2} & \textbf{Stage 3} & \textbf{Stage 4}   \\
    \midrule
    training tokens & 320M & 20M & 1B & 1B \\
    layer-selection score LR & \textemdash & $2\mathrm{e}{-2} \rightarrow 2\mathrm{e}{-3}$ & \textemdash & \textemdash \\
    model weight LR & $1\mathrm{e}{-3} \rightarrow 1\mathrm{e}{-5}$ & $1\mathrm{e}{-4} \rightarrow 1\mathrm{e}{-5}$ & $1\mathrm{e}{-4} \rightarrow 1\mathrm{e}{-5}$ & $1\mathrm{e}{-5}$ \\
    LR scheduler & cosine & WSD& cosine & constant \\
    weight decay & 0 & 0 & 0 & 0 \\
    sequence length & 512 & <4,096 & 512 & 16,384 \\    
    batch size & 32 & 32 & 96 & 128 \\
    warmup steps & 50 & 50 & 50 & 50 \\
    decay steps &19,950 & 50 & 19,950 & 0 \\
    training steps & 20,000 & 250 & 20,000 & 500 \\
    \bottomrule
\end{tabularx}
\end{table*}

\begin{table}[!ht]
    \centering
    \caption{\textbf{Complete Layer Importance Rankings across Backbones and Linear-Attention Variants.} Baseline rankings are adopted from ~\citet{lan2026morphing}. \fbox{Boxed} zero-based indices denote retained full-attention layers ($3{:}1$ ratio), and \textcolor{red}{red} indices indicate selections differing from FlashMorph2.}
    \label{tab:layer_selection_results}
    \setlength{\fboxsep}{1.0pt}
    \setlength{\aboverulesep}{2.0pt}
    \setlength{\belowrulesep}{2.0pt}
    \renewcommand{\arraystretch}{0.85}
    \resizebox{\linewidth}{!}{%
    \begin{tabular}{ll|l}
    \toprule
    \textbf{Attn.} & \textbf{Method} & \textbf{Layer indices (most important $\rightarrow$ least important)} \\
    \midrule
    \multicolumn{3}{l}{\textit{Qwen3-0.6B Backbone}} \\
    \midrule
    \multirow{4}{*}{Lightning}
    & \ourmethod & \fbox{21, 11, 16, 6, 20, 19, 25,} 24, 2, 26, 8, 18, 0, 3, 17, 13, 22, 1, 9, 12, 4, 14, 5, 23, 27, 15, 10, 7 \\
    & FlashMorph & \fbox{\textcolor{red}{1}, 16, 21, 11, 19, \textcolor{red}{24}, \textcolor{red}{0},} 25, 18, 2, 6, 8, 20, 3, 26, 13, 9, 23, 22, 10, 14, 17, 4, 15, 12, 27, 5, 7 \\
    & HALO & \fbox{\textcolor{red}{10}, 21, \textcolor{red}{9}, \textcolor{red}{5}, 11, \textcolor{red}{1}, \textcolor{red}{13},} 16, 25, 12, 19, 24, 6, 18, 15, 8, 2, 26, 14, 23, 0, 27, 20, 7, 17, 22, 3, 4 \\
    & KL-LS & \fbox{21, 16, 19, 20, \textcolor{red}{18}, \textcolor{red}{22}, 25,} 24, 26, 17, 23, 14, 8, 12, 6, 13, 11, 15, 3, 2, 9, 4, 1, 5, 0, 10, 7, 27 \\
    \midrule
    \multirow{4}{*}{GLA}
    & \ourmethod & \fbox{21, 16, 11, 20, 6, 19, 25,} 2, 24, 4, 3, 1, 26, 8, 18, 17, 13, 14, 0, 22, 12, 5, 9, 23, 15, 7, 10, 27 \\
    & FlashMorph & \fbox{21, 16, 11, \textcolor{red}{1}, \textcolor{red}{2}, 6, 25,} 19, 24, 18, 20, 8, 0, 26, 13, 22, 3, 23, 9, 10, 14, 17, 12, 15, 4, 7, 27, 5 \\
    & HALO & \fbox{\textcolor{red}{8}, 21, \textcolor{red}{13}, \textcolor{red}{1}, \textcolor{red}{24}, 6, 25,} 19, 18, 11, 10, 16, 9, 12, 5, 15, 2, 26, 27, 23, 20, 4, 0, 14, 3, 22, 7, 17  \\
    & KL-LS & \fbox{21, 16, 20, 19, \textcolor{red}{18}, \textcolor{red}{24}, \textcolor{red}{8},} 22, 17, 6, 23, 25, 26, 14, 12, 11, 13, 2, 3, 15, 9, 4, 1, 0, 5, 10, 7, 27  \\
    \midrule
    \multirow{4}{*}{GDN}
    & \ourmethod & \fbox{21, 11, 16, 6, 20, 2, 19,} 25, 24, 8, 13, 17, 26, 0, 3, 12, 18, 1, 14, 4, 9, 22, 23, 15, 7, 5, 10, 27 \\
    & FlashMorph & \fbox{\textcolor{red}{1}, 11, 21, 16, 19, \textcolor{red}{18}, \textcolor{red}{24},} 6, 25, 8, 2, 20, 0, 13, 26, 3, 9, 23, 14, 22, 10, 12, 17, 15, 4, 7, 5, 27  \\
    & HALO & \fbox{\textcolor{red}{10}, \textcolor{red}{5}, 21, 6, 19, \textcolor{red}{24}, \textcolor{red}{18},} 11, 9, 13, 12, 25, 8, 2, 16, 26, 1, 4, 15, 0, 17, 7, 23, 14, 27, 3, 22, 20  \\
    & KL-LS & \fbox{21, 16, 19, 20, \textcolor{red}{18}, 6, \textcolor{red}{22},} 24, 25, 11, 8, 14, 26, 12, 23, 17, 13, 3, 2, 9, 1, 4, 15, 5, 0, 10, 27, 7  \\
    \midrule
    \midrule
    \multicolumn{3}{l}{\textit{Qwen3-1.7B Backbone}} \\
    \midrule
    \multirow{5}{*}{Lightning}
    & \ourmethod & \fbox{21, 11, 16, 20, 19, 6, 17,} 2, 13, 0, 18, 8, 24, 25, 3, 26, 12, 1, 14, 9, 22, 15, 4, 10, 5, 23, 7, 27 \\
    & FlashMorph & \fbox{\textcolor{red}{1}, 16, \textcolor{red}{13}, 11, 21, \textcolor{red}{3}, \textcolor{red}{8},} 20, 6, 19, 18, 9, 14, 24, 2, 17, 0, 10, 15, 25, 23, 26, 22, 12, 4, 5, 7, 27 \\
    & HALO & \fbox{\textcolor{red}{3}, \textcolor{red}{14}, \textcolor{red}{9}, \textcolor{red}{2}, 6, 16, 21,} 25, 24, 8, 23, 12, 11, 26, 27, 19, 18, 17, 7, 15, 4, 13, 1, 20, 10, 22, 0, 5 \\
    & KL-LS & \fbox{21, 16, 20, 19, \textcolor{red}{18}, \textcolor{red}{22}, \textcolor{red}{24},} 26, 25, 23, 17, 6, 14, 8, 12, 13, 11, 3, 15, 2, 9, 1, 4, 0, 5, 10, 7, 27  \\
    & PostNAS &  \fbox{\textcolor{red}{0}, 21, \textcolor{red}{25}, 19, 6, 11, \textcolor{red}{9},} 24, 12, 2, 26, 16, 17, 23, 18, 4, 7, 3, 14, 20, 1, 27, 10, 13, 8, 22, 15, 5 \\
    \midrule
    \multirow{5}{*}{GLA}
    & \ourmethod & \fbox{21, 16, 11, 20, 19, 6, 2,} 17, 1, 13, 0, 8, 3, 25, 12, 14, 24, 18, 5, 4, 9, 26, 22, 7, 15, 23, 10, 27 \\
    & FlashMorph & \fbox{\textcolor{red}{1}, 11, 21, \textcolor{red}{3}, \textcolor{red}{13}, \textcolor{red}{14}, 16,} 19, 9, 20, 6, 18, 2, 0, 24, 8, 10, 17, 25, 15, 23, 26, 12, 22, 4, 7, 27, 5  \\
    & HALO & \fbox{\textcolor{red}{3}, \textcolor{red}{14}, 6, \textcolor{red}{8}, \textcolor{red}{4}, \textcolor{red}{25}, 11,} 21, 16, 2, 24, 18, 26, 17, 19, 23, 1, 27, 12, 0, 15, 13, 7, 20, 9, 22, 10, 5  \\
    & KL-LS & \fbox{21, 16, 20, 19, \textcolor{red}{18}, \textcolor{red}{24}, \textcolor{red}{22},} 6, 8, 26, 11, 25, 23, 14, 12, 17, 3, 13, 2, 1, 9, 4, 15, 0, 5, 10, 27, 7 \\
    & PostNAS &  \fbox{\textcolor{red}{0}, 21, \textcolor{red}{25}, 19, 6, 11, \textcolor{red}{9},} 24, 12, 2, 26, 16, 17, 23, 18, 4, 7, 3, 14, 20, 1, 27, 10, 13, 8, 22, 15, 5  \\
    \midrule
    \multirow{5}{*}{GDN}
    & \ourmethod & \fbox{21, 11, 16, 20, 6, 17, 19,} 2, 13, 8, 12, 14, 3, 0, 1, 24, 25, 18, 26, 9, 22, 4, 23, 15, 7, 10, 5, 27 \\
    & FlashMorph &  \fbox{\textcolor{red}{1}, 11, \textcolor{red}{13}, 21, 16, \textcolor{red}{14}, 6,} 10, 20, 19, 2, 18, 8, 24, 0, 3, 9, 25, 15, 22, 17, 26, 23, 12, 4, 27, 7, 5  \\
    & HALO & \fbox{\textcolor{red}{3}, \textcolor{red}{2}, \textcolor{red}{25}, 21, 11, \textcolor{red}{14}, 6,} 4, 12, 8, 16, 18, 24, 17, 19, 23, 7, 26, 27, 9, 1, 20, 22, 13, 0, 15, 5, 10 \\
    & KL-LS & \fbox{21, 16, 20, 19, 6, 11, \textcolor{red}{18},} 24, 22, 8, 25, 26, 12, 14, 23, 13, 2, 3, 17, 1, 9, 4, 15, 5, 10, 0, 27, 7 \\
    & PostNAS &  \fbox{\textcolor{red}{0}, 21, \textcolor{red}{25}, 19, 6, 11, \textcolor{red}{9},} 24, 12, 2, 26, 16, 17, 23, 18, 4, 7, 3, 14, 20, 1, 27, 10, 13, 8, 22, 15, 5  \\
    \midrule
    \midrule
    \multicolumn{3}{l}{\textit{Qwen3-8B Backbone}} \\
    \midrule
    \multirow{6}{*}{Lightning}
    & \ourmethod & \makecell[l]{\fbox{13, 29, 24, 20, 33, 15, 22, 17, 9,} 3, 19, 23, 0, 18, 16, 5, 32, 21, 1, 2, 34, 11, 4, 7, 8, 6, 26, 14, \\12, 10, 31, 27, 28, 25, 30, 35} \\
    & FlashMorph & \makecell[l]{\fbox{\textcolor{red}{1}, \textcolor{red}{0}, \textcolor{red}{7}, 22, 29, 13, \textcolor{red}{3}, 15, 24,} 20, 9, 19, 33, 34, 12, 2, 17, 8, 32, 16, 14, 23, 21, 35, 18, 5, 31, \\11, 10, 4, 28, 26, 30, 6, 25, 27} \\
    & HALO & \makecell[l]{\fbox{\textcolor{red}{7}, \textcolor{red}{6}, \textcolor{red}{0}, 24, \textcolor{red}{8}, 33, \textcolor{red}{1}, \textcolor{red}{12}, \textcolor{red}{34},} 22, 2, 15, 26, 9, 20, 16, 29, 31, 21, 4, 17, 30, 5, 35, 11, 32, 3, 25, \\27, 14, 18, 19, 28, 13, 23, 10} \\
    \midrule
    \midrule
    \multicolumn{3}{l}{\textit{Qwen3-30B-A3B Backbone}} \\
    \midrule
    \multirow{6}{*}{Lightning}
    & \ourmethod & \makecell[l]{\fbox{42, 45, 0, 1, 7, 22, 6, 43, 39, 5, 11, 2,} 38, 27, 10, 26, 24, 34, 8, 13, 18, 30, 9, 15, 23, 12, 37, \\17, 14, 3, 4,16, 25, 20, 46,41, 36, 21, 44, 31, 33, 19, 28, 29, 40, 35, 47, 32} \\
    & FlashMorph & \makecell[l]{\fbox{\textcolor{red}{37}, 2, \textcolor{red}{4}, \textcolor{red}{3}, 1, \textcolor{red}{38}, 5, 0, 6, 42, \textcolor{red}{8}, \textcolor{red}{26},} 13, 22, 18, 21, 11, 9, 14, 20, 15, 25, 36, 24, 10, 41, 12, \\34, 47, 17, 23, 30, 7, 45, 39, 16, 19, 29, 35, 43, 33, 46, 27, 44, 40, 32, 28, 31} \\
    & HALO & \makecell[l]{\fbox{5, 45, 42, \textcolor{red}{8}, \textcolor{red}{12}, 43, \textcolor{red}{36}, \textcolor{red}{19}, \textcolor{red}{34}, \textcolor{red}{41}, \textcolor{red}{18}, 22,} 46, 27, 37, 4, 31, 39, 13, 9, 35, 10, 24, 7, 33, 25,\\47, 23, 28, 17, 21, 11, 14, 0, 30, 15, 20, 26, 40, 3, 1, 6, 32, 38, 2, 29, 16, 44 } \\
    \bottomrule
    \end{tabular}
    }
\end{table}

\clearpage
\section{Evaluation}
\label{app:eval}

\textbf{Long-context Capabilities.} We evaluate long-context retrieval and understanding on RULER~\citep{hsieh2024ruler} and LongBench~\citep{bai2024longbench}. For RULER, we evaluate across all 13 tasks grouped into four categories: Needle-in-a-Haystack (NIAH), including Single-NIAH (SG1--SG3), Multi-keys NIAH (MK1--MK3), Multi-queries NIAH (MQ), and Multi-values NIAH (MV); Variable Tracking (VT); Aggregation (Agg.), including Common Words Extraction (CWE) and Frequent Words Extraction (FWE); and Question Answering, including QA1~\citep{rajpurkar2016squad}, and QA2~\citep{yang2018hotpotqa}. For LongBench, we evaluate on 14 tasks spanning five categories: Single-Document QA, including NarrativeQA (NQA;~\citealt{kovcisky2018narrativeqa}), Qasper (QQA;~\citealt{dasigi2021dataset}), and MultiFieldQA (MFQ;~\citealt{bai2024longbench});  Multi-Document QA, including HotpotQA (HQA;~\citealt{yang2018hotpotqa}), 2WikiMultihopQA (2WM;~\citealt{ho2020constructing}), and MuSiQue (Mus;~\citealt{trivedi2021musique}); Summarization, including GovReport (GvR;~\citealt{huang2021efficient}), QMSum (QMS;~\citealt{zhong2021qmsum}), and MultiNews (MNs;~\citealt{fabbri2019multi}); Few-shot Learning (Few-shot), including TREC (TRC;~\citealt{li2002learning}), TriviaQA (TQA;~\citealt{joshi2017triviaqa}), and SAMSum (SSM;~\citealt{gliwa-etal-2019-samsum});  and Code Completion (Code), including Long Code Completion (LCC;~\citealt{guo2023longcoder}) and RepoBench-P (RBP;~\citealt{liu2023repobench}).

\textbf{Recall-intensive Tasks.} We evaluate real-world recall-intensive tasks across six datasets: FDA~\citep{arora2023language} for PDF key-value retrieval, SWDE~\citep{lockard2019openceres} for structured HTML relation extraction, and four question-answering datasets including SQuAD (SQD;~\citealt{rajpurkar-etal-2018-know}), Natural Questions (NQ;~\citealt{kwiatkowski2019natural}), TriviaQA (TQA;~\citealt{joshi2017triviaqa}), and DROP~\citep{dua2019drop}.

\textbf{Language Modeling and Commonsense Reasoning.} Following previous work~\citep{gu2023mamba,yang2025gated}, we evaluate our method on language modeling perplexity and zero-shot commonsense reasoning benchmarks: WikiText (Wiki.;~\citealt{merity2016pointer}), LAMBDA (LMB.;~\citealt{paperno-etal-2016-lambada}), ARC-easy (ARC-e) and ARC-challenge (ARC-c)~\citep{clark2018think}, PIQA~\citep{bisk2020piqa}, HellaSwag (Hella.;~\citealt{zellers2019hellaswag}), WinoGrande (Wino.;~\citealt{sakaguchi2021winogrande}), SIQA~\citep{sap-etal-2019-social}, and BoolQ~\citep{clark-etal-2019-boolq}.

\section{Detailed Benchmark Results}
\label{app:results}

Tables~\ref{tab:ruler}, \ref{tab:longbench}, and \ref{tab:zero_shot} present the detailed per-task results on RULER, LongBench, and language modeling and zero-shot commonsense reasoning benchmarks, respectively. In addition to the Qwen3-0.6B and Qwen3-1.7B configurations reported in the main text, we further scale our evaluation to larger dense and MoE backbones, including Qwen3-8B and Qwen3-30B-A3B with Lightning Attention. Across these larger model scales, FlashMorph2 continues to consistently outperform competing layer-selection baselines on both RULER and LongBench, while preserving strong language modeling perplexity and zero-shot commonsense reasoning accuracy.

\begin{table*}[t]
    \centering
    \caption{\textbf{Detailed Results on RULER.} Per-task scores and average across all 13 RULER tasks. For each backbone and linear-attention variant, the \colorbox{cyan!25}{best} and \colorbox{cyan!10}{second-best} scores among layer-selection methods are highlighted.}
    \label{tab:ruler}
    \small
    \setlength{\tabcolsep}{2.5pt}
    \renewcommand{\arraystretch}{0.92}
    \begin{tabular}{ll|ccccccccccccc|c}
    \toprule
    \textbf{Attn.} & \textbf{Method} & \textbf{SG1} & \textbf{SG2} & \textbf{SG3} & \textbf{MK1} & \textbf{MK2} & \textbf{MK3} & \textbf{MQ} & \textbf{MV} & \textbf{VT} & \textbf{CWE} & \textbf{FWE} & \textbf{QA1} & \textbf{QA2} & \textbf{Avg} \\
    \midrule
    
    \multicolumn{16}{l}{\textit{0.6B Backbone}} \\
    \rowcolor{gray!15} 
    Full & Qwen3 & 100.00 & 100.00 & 99.60 & 99.20 & 99.00 & 63.60 & 97.70 & 95.90 & 95.32 & 52.96 & 58.60 & 41.32 & 25.80 & 79.15 \\
    \midrule
    \multirow{5}{*}{Lightning}
    & Uniform & \cellcolor{cyan!10}99.80 & \cellcolor{cyan!10}99.80 & \cellcolor{cyan!10}95.60 & 65.80 & 15.40 & 0.40 & 53.25 & 58.95 & 1.56 & 46.62 & \cellcolor{cyan!10}56.67 & 36.58 & 23.40 & 50.29 \\
    & KL-LS & \cellcolor{cyan!25}100.00 & \cellcolor{cyan!25}100.00 & 73.60 & 42.20 & 30.80 & 0.20 & 48.35 & 40.75 & 2.20 & 18.52 & 50.40 & 37.32 & 26.00 & 43.87 \\
    & HALO & 83.80 & \cellcolor{cyan!25}100.00 & \cellcolor{cyan!25}97.40 & 62.00 & 13.20 & 1.00 & 60.05 & 58.60 & 1.20 & 48.04 & 52.93 & 38.62 & 26.60 & 49.50 \\
    & FlashMorph & \cellcolor{cyan!25}100.00 & \cellcolor{cyan!25}100.00 & 94.20 & \cellcolor{cyan!10}79.00 & \cellcolor{cyan!25}43.40 & \cellcolor{cyan!10}1.40 & \cellcolor{cyan!10}64.50 & \cellcolor{cyan!10}66.00 & \cellcolor{cyan!10}10.84 & \cellcolor{cyan!25}54.80 & 55.27 & \cellcolor{cyan!10}42.25 & \cellcolor{cyan!25}28.40 & \cellcolor{cyan!10}56.93 \\
    & \ourmethod & \cellcolor{cyan!25}100.00 & \cellcolor{cyan!25}100.00 & 88.00 & \cellcolor{cyan!25}80.00 & \cellcolor{cyan!10}31.60 & \cellcolor{cyan!25}3.40 & \cellcolor{cyan!25}65.30 & \cellcolor{cyan!25}79.85 & \cellcolor{cyan!25}17.24 & \cellcolor{cyan!10}51.80 & \cellcolor{cyan!25}65.13 & \cellcolor{cyan!25}45.18 & \cellcolor{cyan!10}27.80 & \cellcolor{cyan!25}58.10 \\

    \midrule
    \multirow{5}{*}{GLA}
    & Uniform & \cellcolor{cyan!25}100.00 & \cellcolor{cyan!25}100.00 & 88.40 & 74.80 & 26.20 & 1.00 & 52.95 & 53.90 & 4.44 & 50.48 & 47.13 & 32.87 & 21.40 & 50.27 \\
    & KL-LS & \cellcolor{cyan!25}100.00 & \cellcolor{cyan!25}100.00 & \cellcolor{cyan!25}98.20 & 81.20 & \cellcolor{cyan!10}32.00 & 2.20 & 68.45 & 61.55 & 1.80 & 37.06 & \cellcolor{cyan!25}49.87 & 35.90 & 24.00 & 53.25 \\
    & HALO & \cellcolor{cyan!25}100.00 & \cellcolor{cyan!25}100.00 & \cellcolor{cyan!10}97.80 & 77.60 & 15.40 & 0.80 & 63.85 & 71.30 & 2.08 & \cellcolor{cyan!25}58.54 & 36.20 & 36.97 & 23.00 & 52.58 \\
    & FlashMorph & \cellcolor{cyan!10}99.80 & \cellcolor{cyan!25}100.00 & 97.60 & \cellcolor{cyan!10}88.00 & 18.60 & \cellcolor{cyan!10}2.80 & \cellcolor{cyan!10}71.65 & \cellcolor{cyan!10}73.50 & \cellcolor{cyan!10}5.52 & 53.06 & \cellcolor{cyan!10}49.47 & \cellcolor{cyan!10}39.92 & \cellcolor{cyan!10}25.20 & \cellcolor{cyan!10}55.78 \\
    & \ourmethod & \cellcolor{cyan!25}100.00 & \cellcolor{cyan!25}100.00 & 89.40 & \cellcolor{cyan!25}92.40 & \cellcolor{cyan!25}61.60 & \cellcolor{cyan!25}4.20 & \cellcolor{cyan!25}83.05 & \cellcolor{cyan!25}78.65 & \cellcolor{cyan!25}7.20 & \cellcolor{cyan!10}57.84 & 47.40 & \cellcolor{cyan!25}44.25 & \cellcolor{cyan!25}29.00 & \cellcolor{cyan!25}61.15 \\

    \midrule
    \multirow{5}{*}{GDN}
    & Uniform & \cellcolor{cyan!10}99.60 & \cellcolor{cyan!10}99.80 & \cellcolor{cyan!10}94.60 & 70.80 & 30.20 & 1.40 & 56.30 & 60.00 & 3.76 & 48.38 & 52.07 & 40.32 & 26.20 & 52.57 \\
    & KL-LS & \cellcolor{cyan!25}100.00 & \cellcolor{cyan!25}100.00 & 92.80 & 81.80 & \cellcolor{cyan!10}56.00 & 3.00 & \cellcolor{cyan!10}72.20 & 63.00 & 2.96 & 41.14 & 56.80 & 42.12 & 27.40 & 56.86 \\
    & HALO & \cellcolor{cyan!25}100.00 & \cellcolor{cyan!25}100.00 & 85.00 & 69.20 & 25.80 & 2.00 & 54.40 & 45.20 & \cellcolor{cyan!25}20.28 & 35.36 & 54.26 & 39.67 & 27.20 & 50.64 \\
    & FlashMorph & \cellcolor{cyan!25}100.00 & \cellcolor{cyan!25}100.00 & 91.20 & \cellcolor{cyan!10}89.00 & 52.00 & \cellcolor{cyan!10}3.20 & \cellcolor{cyan!25}75.15 & \cellcolor{cyan!10}69.30 & 15.60 & \cellcolor{cyan!10}55.08 & \cellcolor{cyan!10}58.00 & \cellcolor{cyan!25}46.10 & \cellcolor{cyan!25}29.20 & \cellcolor{cyan!10}60.29 \\
    & \ourmethod & \cellcolor{cyan!25}100.00 & \cellcolor{cyan!25}100.00 & \cellcolor{cyan!25}98.80 & \cellcolor{cyan!25}95.20 & \cellcolor{cyan!25}79.60 & \cellcolor{cyan!25}4.40 & 63.80 & \cellcolor{cyan!25}72.85 & \cellcolor{cyan!10}18.68 & \cellcolor{cyan!25}60.88 & \cellcolor{cyan!25}60.20 & \cellcolor{cyan!10}45.18 & \cellcolor{cyan!10}27.80 & \cellcolor{cyan!25}63.65 \\

    \midrule
    \midrule
    
    \multicolumn{16}{l}{\textit{1.7B Backbone}} \\
    \rowcolor{gray!15} 
    Full & Qwen3 & 100.00 & 100.00 & 99.80 & 99.80 & 98.20 & 98.40 & 99.90 & 99.95 & 95.72 & 88.62 & 66.33 & 51.88 & 41.00 & 87.66  \\

    \midrule
    \multirow{6}{*}{Lightning}
    & Uniform & \cellcolor{cyan!25}100.00 & \cellcolor{cyan!25}100.00 & 87.80 & 73.60 & \cellcolor{cyan!10}67.60 & \cellcolor{cyan!10}2.00 & 75.75 & 64.10 & 29.16 & 42.20 & 61.73 & 46.45 & 35.00 & 60.41 \\
    & PostNAS & \cellcolor{cyan!25}100.00 & \cellcolor{cyan!10}99.80 & \cellcolor{cyan!10}97.40 & 76.40 & 58.20 & 1.80 & 55.15 & 56.65 & 23.16 & \cellcolor{cyan!25}63.36 & 66.53 & 48.15 & 37.60 & 60.32 \\
    & KL-LS & \cellcolor{cyan!25}100.00 & \cellcolor{cyan!25}100.00 & 70.80 & 49.60 & 42.00 & 0.60 & 44.55 & 48.05 & 0.24 & 42.02 & 46.00 & 42.05 & 32.40 & 47.56 \\
    & HALO & \cellcolor{cyan!25}100.00 & \cellcolor{cyan!25}100.00 & 93.00 & \cellcolor{cyan!10}84.60 & 45.00 & 1.80 & 77.30 & 81.40 & 29.48 & \cellcolor{cyan!10}59.94 & \cellcolor{cyan!25}67.27 & 52.02 & 35.80 & 63.66 \\
    & FlashMorph & \cellcolor{cyan!10}99.80 & \cellcolor{cyan!25}100.00 & \cellcolor{cyan!25}98.00 & \cellcolor{cyan!25}86.80 & 56.20 & \cellcolor{cyan!10}2.00 & \cellcolor{cyan!25}86.80 & \cellcolor{cyan!25}91.15 & \cellcolor{cyan!10}31.64 & 58.78 & 61.33 & \cellcolor{cyan!10}52.22 & \cellcolor{cyan!10}37.80 & \cellcolor{cyan!10}66.35 \\
    & \ourmethod & \cellcolor{cyan!25}100.00 & \cellcolor{cyan!25}100.00 & 96.80 & 80.40 & \cellcolor{cyan!25}87.00 & \cellcolor{cyan!25}7.80 & \cellcolor{cyan!10}81.85 & \cellcolor{cyan!10}85.80 & \cellcolor{cyan!25}41.24 & 58.62 & \cellcolor{cyan!10}67.20 & \cellcolor{cyan!25}56.08 & \cellcolor{cyan!25}39.40 & \cellcolor{cyan!25}69.40 \\

    \midrule
    \multirow{6}{*}{GLA}
    & Uniform & \cellcolor{cyan!25}100.00 & \cellcolor{cyan!25}100.00 & 86.40 & 83.20 & \cellcolor{cyan!10}65.00 & 2.00 & 68.45 & 77.85 & 20.56 & 58.28 & 51.67 & 48.57 & 35.40 & 61.34 \\
    & PostNAS & \cellcolor{cyan!10}99.60 & \cellcolor{cyan!25}100.00 & \cellcolor{cyan!25}99.20 & \cellcolor{cyan!10}92.00 & 48.40 & 3.60 & 70.20 & 86.60 & 19.76 & 56.04 & 47.13 & \cellcolor{cyan!10}52.25 & \cellcolor{cyan!25}39.60 & 62.64 \\
    & KL-LS & \cellcolor{cyan!25}100.00 & \cellcolor{cyan!25}100.00 & 65.00 & 43.80 & 48.00 & 2.40 & 43.15 & 69.15 & 11.60 & 54.28 & 46.67 & 41.98 & 35.20 & 50.86 \\
    & HALO & \cellcolor{cyan!25}100.00 & \cellcolor{cyan!10}99.60 & 95.60 & 68.40 & 25.80 & 0.80 & 58.00 & 67.40 & \cellcolor{cyan!10}25.84 & \cellcolor{cyan!25}63.52 & \cellcolor{cyan!10}60.87 & 45.17 & 29.60 & 56.97 \\
    & FlashMorph & \cellcolor{cyan!25}100.00 & \cellcolor{cyan!10}99.60 & 95.60 & 87.00 & \cellcolor{cyan!25}80.60 & \cellcolor{cyan!10}5.00 & \cellcolor{cyan!25}83.20 & \cellcolor{cyan!10}92.45 & \cellcolor{cyan!25}30.52 & 57.50 & 55.47 & 49.65 & 36.00 & \cellcolor{cyan!10}67.12 \\
    & \ourmethod & \cellcolor{cyan!25}100.00 & \cellcolor{cyan!25}100.00 & \cellcolor{cyan!10}98.40 & \cellcolor{cyan!25}94.40 & 63.00 & \cellcolor{cyan!25}8.00 & \cellcolor{cyan!10}76.75 & \cellcolor{cyan!25}92.60 & 22.28 & \cellcolor{cyan!10}61.22 & \cellcolor{cyan!25}62.40 & \cellcolor{cyan!25}58.22 & \cellcolor{cyan!10}37.80 & \cellcolor{cyan!25}67.31 \\

    \midrule
    \multirow{6}{*}{GDN}
    & Uniform & \cellcolor{cyan!25}100.00 & \cellcolor{cyan!25}100.00 & 97.20 & 75.80 & 64.40 & 3.20 & 76.45 & 87.45 & 63.32 & 51.70 & 67.13 & 51.65 & 39.20 & 67.50 \\
    & PostNAS & \cellcolor{cyan!25}100.00 & \cellcolor{cyan!25}100.00 & \cellcolor{cyan!25}100.00 & 88.80 & \cellcolor{cyan!10}75.00 & 6.00 & 74.35 & 89.45 & 53.16 & \cellcolor{cyan!10}61.54 & 68.40 & 53.25 & 36.60 & 69.73 \\
    & KL-LS & \cellcolor{cyan!25}100.00 & \cellcolor{cyan!25}100.00 & \cellcolor{cyan!10}99.00 & 89.40 & 60.00 & \cellcolor{cyan!10}9.60 & 87.55 & 92.10 & \cellcolor{cyan!25}68.12 & 61.40 & \cellcolor{cyan!10}69.47 & 55.22 & 40.60 & 71.73 \\
    & HALO & \cellcolor{cyan!25}100.00 & \cellcolor{cyan!25}100.00 & 98.60 & 88.60 & 50.20 & 2.40 & 81.40 & 88.25 & 62.68 & 58.88 & 67.73 & 54.88 & 39.00 & 68.66 \\
    & FlashMorph & \cellcolor{cyan!25}100.00 & \cellcolor{cyan!10}99.80 & \cellcolor{cyan!25}100.00 & \cellcolor{cyan!25}90.60 & 66.80 & 8.60 & \cellcolor{cyan!25}92.50 & \cellcolor{cyan!25}96.05 & \cellcolor{cyan!10}66.76 & 60.84 & \cellcolor{cyan!25}70.93 & \cellcolor{cyan!10}55.25 & \cellcolor{cyan!10}41.80 & \cellcolor{cyan!10}73.07 \\
    & \ourmethod & \cellcolor{cyan!25}100.00 & \cellcolor{cyan!25}100.00 & 98.60 & \cellcolor{cyan!10}90.00 & \cellcolor{cyan!25}91.80 & \cellcolor{cyan!25}12.80 & \cellcolor{cyan!10}90.30 & \cellcolor{cyan!10}95.15 & 58.80 & \cellcolor{cyan!25}63.90 & 68.67 & \cellcolor{cyan!25}55.52 & \cellcolor{cyan!25}42.40 & \cellcolor{cyan!25}74.46 \\

    \midrule
    \midrule
    \multicolumn{16}{l}{\textit{8B Backbone}} \\
    \rowcolor{gray!15} 
    Full & Qwen3 & 100.00 & 100.00 & 100.00 & 100.00 & 100.00 & 100.00 & 100.00 & 99.85 & 99.96 & 99.00 & 93.73 & 71.32 & 56.60 & 93.88 \\
    \midrule
    \multirow{4}{*}{Lightning}
    & Uniform & \cellcolor{cyan!25}100.00 & \cellcolor{cyan!10}99.80 & \cellcolor{cyan!10}99.80 & 87.80 & \cellcolor{cyan!10}76.80 & 7.20 & 90.85 & 91.25 & \cellcolor{cyan!10}65.20 & 61.22 & \cellcolor{cyan!25}81.13 & \cellcolor{cyan!10}58.50 & 43.00 & 74.04 \\
    & HALO & \cellcolor{cyan!25}100.00 & \cellcolor{cyan!25}100.00 & 96.40 & 56.40 & 22.40 & 1.80 & 86.40 & 90.55 & 55.32 & \cellcolor{cyan!25}67.86 & 55.07 & 51.35 & 38.20 & 63.21 \\
    & FlashMorph & \cellcolor{cyan!25}100.00 & \cellcolor{cyan!25}100.00 & \cellcolor{cyan!10}99.80 & \cellcolor{cyan!25}90.20 & \cellcolor{cyan!25}80.80 & \cellcolor{cyan!10}17.80 & \cellcolor{cyan!25}95.95 & \cellcolor{cyan!25}97.75 & 51.00 & 58.66 & \cellcolor{cyan!10}70.53 & \cellcolor{cyan!25}60.98 & \cellcolor{cyan!10}43.40 & \cellcolor{cyan!10}74.38 \\
    & \ourmethod & \cellcolor{cyan!25}100.00 & \cellcolor{cyan!25}100.00 & \cellcolor{cyan!25}100.00 & \cellcolor{cyan!10}90.00 & 71.20 & \cellcolor{cyan!25}28.60 & \cellcolor{cyan!10}95.30 & \cellcolor{cyan!10}95.05 & \cellcolor{cyan!25}83.44 & \cellcolor{cyan!10}64.44 & 68.93 & 54.68 & \cellcolor{cyan!25}44.40 & \cellcolor{cyan!25}76.62 \\
    \midrule
    \midrule
    \multicolumn{16}{l}{\textit{30B-A3B Backbone}} \\
    \rowcolor{gray!15} 
    Full & Qwen3 & 100.00 & 100.00 & 100.00 & 100.00 & 100.00 & 100.00 & 91.80 & 31.80 & 99.96 & 59.28 & 67.00 & 58.58 & 39.80 & 80.63 \\
    \midrule
    \multirow{4}{*}{Lightning}
    & Uniform & \cellcolor{cyan!10}99.60 & \cellcolor{cyan!25}100.00 & 97.00 & 58.80 & 16.20 & 2.20 & 85.55 & 87.45 & 32.76 & 57.46 & 61.47 & 43.05 & \cellcolor{cyan!25}40.00 & 60.12 \\
    & HALO & \cellcolor{cyan!25}100.00 & \cellcolor{cyan!25}100.00 & \cellcolor{cyan!10}98.80 & 70.20 & 15.00 & 1.40 & 81.70 & 93.15 & \cellcolor{cyan!10}36.08 & \cellcolor{cyan!10}63.10 & 57.00 & 34.82 & \cellcolor{cyan!10}38.00 & 60.71 \\
    & FlashMorph & \cellcolor{cyan!25}100.00 & \cellcolor{cyan!25}100.00 & \cellcolor{cyan!25}99.20 & \cellcolor{cyan!10}79.20 & \cellcolor{cyan!10}53.60 & \cellcolor{cyan!10}6.20 & \cellcolor{cyan!25}97.75 & \cellcolor{cyan!25}97.40 & 19.04 & 61.76 & \cellcolor{cyan!10}69.40 & \cellcolor{cyan!25}45.77 & 33.80 & \cellcolor{cyan!10}66.39 \\
    & \ourmethod & \cellcolor{cyan!25}100.00 & \cellcolor{cyan!25}100.00 & 90.80 & \cellcolor{cyan!25}80.20 & \cellcolor{cyan!25}54.80 & \cellcolor{cyan!25}7.20 & \cellcolor{cyan!10}96.35 & \cellcolor{cyan!10}96.60 & \cellcolor{cyan!25}44.96 & \cellcolor{cyan!25}65.78 & \cellcolor{cyan!25}71.47 & \cellcolor{cyan!10}45.73 & 33.20 & \cellcolor{cyan!25}68.24 \\
    \bottomrule
    \end{tabular}
\end{table*}

\begin{table*}[t]
    \centering
    \caption{\textbf{Detailed Results on LongBench.} Per-task scores and average across 14 LongBench tasks spanning five categories. For each backbone and linear-attention variant, the \colorbox{cyan!25}{best} and \colorbox{cyan!10}{second-best} scores among layer-selection methods are highlighted.}
    \label{tab:longbench}
    \small
    \setlength{\tabcolsep}{2.7pt}
    \renewcommand{\arraystretch}{0.92}
    \begin{tabular}{ll|ccc|ccc|ccc|ccc|cc|c}
    \toprule
    \multirow{2}{*}{\textbf{Attn.}}
    & \multirow{2}{*}{\textbf{Method}} 
    & \multicolumn{3}{c|}{\textbf{Single-Doc QA}} & \multicolumn{3}{c|}{\textbf{Multi-Doc QA}} & 
    \multicolumn{3}{c|}{\textbf{Summarization}} & \multicolumn{3}{c|}{\textbf{Few-shot}} & \multicolumn{2}{c|}{\textbf{Code}} &  \multirow{2}{*}{\textbf{Avg}} \\
    & & NQA & QQA & MFQ & HQA & 2WM & Mus & GvR & QMS & MNs & TRC & TQA & SSM & LCC & RBP & \\
    \midrule
    
    \multicolumn{17}{l}{\textit{0.6B Backbone}} \\
    \rowcolor{gray!15}
    Full & Qwen3 & 10.04 & 14.12 & 39.49 & 26.25 & 25.91 & 10.41 & 25.05 & 19.10 & 14.03 & 70.50 & 73.49 & 38.34 & 31.31 & 35.21 & 30.95 \\

    \midrule
    \multirow{5}{*}{Lightning}
    & Uniform & \cellcolor{cyan!10}14.34 & 17.78 & 27.59 & 20.21 & 20.80 & 8.38 & 15.87 & 20.36 & 17.52 & \cellcolor{cyan!25}56.00 & \cellcolor{cyan!10}55.80 & \cellcolor{cyan!25}33.95 & 30.14 & 36.42 & 26.80 \\
    & KL-LS & 11.50 & 18.76 & 25.30 & \cellcolor{cyan!10}24.76 & \cellcolor{cyan!25}26.37 & \cellcolor{cyan!25}9.32 & 17.15 & 20.16 & 18.50 & 48.50 & 49.03 & 32.44 & \cellcolor{cyan!10}36.22 & \cellcolor{cyan!10}41.20 & 27.09 \\
    & HALO & 10.33 & 19.44 & 27.39 & 22.66 & 24.81 & 7.25 & 13.70 & \cellcolor{cyan!10}20.58 & 17.05 & 50.00 & 53.28 & 30.87 & 32.19 & 37.11 & 26.19 \\
    & FlashMorph & \cellcolor{cyan!25}15.05 & \cellcolor{cyan!25}23.34 & \cellcolor{cyan!25}31.69 & \cellcolor{cyan!25}24.95 & 25.29 & 7.24 & \cellcolor{cyan!10}18.30 & 20.50 & \cellcolor{cyan!25}19.86 & 51.00 & 55.01 & 33.27 & 33.46 & 40.01 & \cellcolor{cyan!10}28.50 \\
    & \ourmethod & 14.05 & \cellcolor{cyan!10}23.09 & \cellcolor{cyan!10}31.56 & 23.72 & \cellcolor{cyan!10}25.44 & \cellcolor{cyan!10}9.02 & \cellcolor{cyan!25}19.46 & \cellcolor{cyan!25}20.89 & \cellcolor{cyan!10}19.06 & \cellcolor{cyan!10}52.50 & \cellcolor{cyan!25}56.17 & \cellcolor{cyan!10}33.36 & \cellcolor{cyan!25}36.71 & \cellcolor{cyan!25}43.23 & \cellcolor{cyan!25}29.16 \\

    \midrule
    \multirow{5}{*}{GLA}
    & Uniform & 11.70 & 19.82 & 27.78 & 16.78 & 22.31 & 7.73 & 15.30 & 18.50 & 16.67 & \cellcolor{cyan!25}58.00 & 49.11 & 33.52 & 27.43 & 31.69 & 25.45 \\
    & KL-LS & \cellcolor{cyan!10}14.09 & \cellcolor{cyan!10}24.84 & 26.66 & 19.46 & 24.18 & \cellcolor{cyan!25}9.28 & \cellcolor{cyan!25}20.02 & 18.98 & \cellcolor{cyan!10}18.58 & \cellcolor{cyan!10}52.50 & \cellcolor{cyan!10}56.56 & \cellcolor{cyan!10}34.66 & \cellcolor{cyan!10}30.25 & 31.32 & \cellcolor{cyan!10}27.24 \\
    & HALO & 11.96 & 19.28 & 28.86 & \cellcolor{cyan!25}24.68 & \cellcolor{cyan!10}24.24 & 8.76 & 15.70 & \cellcolor{cyan!25}20.54 & 15.87 & 48.50 & \cellcolor{cyan!25}56.62 & \cellcolor{cyan!25}35.38 & 26.17 & \cellcolor{cyan!10}31.83 & 26.31 \\
    & FlashMorph & 12.95 & \cellcolor{cyan!25}26.31 & \cellcolor{cyan!25}31.66 & \cellcolor{cyan!10}23.87 & 24.08 & \cellcolor{cyan!10}9.02 & 16.53 & 20.29 & 18.37 & 46.50 & 51.67 & 33.00 & 28.70 & 30.58 & 26.68 \\
    & \ourmethod & \cellcolor{cyan!25}15.23 & 23.66 & \cellcolor{cyan!10}31.59 & 21.31 & \cellcolor{cyan!25}24.31 & 8.18 & \cellcolor{cyan!10}18.92 & \cellcolor{cyan!10}20.45 & \cellcolor{cyan!25}19.02 & 49.50 & 55.75 & 34.01 & \cellcolor{cyan!25}30.65 & \cellcolor{cyan!25}33.31 & \cellcolor{cyan!25}27.57 \\

    \midrule
    \multirow{5}{*}{GDN}
    & Uniform & 14.49 & 20.92 & 28.15 & 21.79 & 18.33 & \cellcolor{cyan!10}9.09 & 16.61 & \cellcolor{cyan!10}21.39 & 15.66 & 60.00 & 52.86 & \cellcolor{cyan!10}34.42 & 30.39 & 39.99 & 27.43 \\
    & KL-LS & 13.95 & 21.14 & 27.59 & 23.16 & 20.13 & 8.45 & \cellcolor{cyan!25}19.84 & 21.22 & \cellcolor{cyan!25}20.00 & \cellcolor{cyan!25}63.00 & \cellcolor{cyan!10}53.66 & 33.16 & \cellcolor{cyan!25}35.42 & 42.28 & 28.79 \\
    & HALO & 14.19 & 22.03 & 26.13 & 20.48 & 21.66 & 6.58 & 18.95 & 20.55 & 15.93 & 55.50 & 49.76 & 34.16 & 33.23 & 40.70 & 27.13 \\
    & FlashMorph & \cellcolor{cyan!10}14.74 & \cellcolor{cyan!10}23.05 & \cellcolor{cyan!10}32.31 & \cellcolor{cyan!10}24.22 & \cellcolor{cyan!10}24.19 & 8.31 & \cellcolor{cyan!10}19.24 & 20.70 & 19.10 & 58.50 & \cellcolor{cyan!25}54.45 & 33.39 & 33.33 & \cellcolor{cyan!25}42.55 & \cellcolor{cyan!10}29.15 \\
    & \ourmethod & \cellcolor{cyan!25}15.13 & \cellcolor{cyan!25}23.42 & \cellcolor{cyan!25}34.07 & \cellcolor{cyan!25}24.38 & \cellcolor{cyan!25}25.17 & \cellcolor{cyan!25}10.57 & 19.03 & \cellcolor{cyan!25}22.11 & \cellcolor{cyan!10}19.31 & \cellcolor{cyan!10}62.00 & 51.17 & \cellcolor{cyan!25}34.71 & \cellcolor{cyan!10}33.34 & \cellcolor{cyan!10}42.41 & \cellcolor{cyan!25}29.77 \\

    \midrule
    \midrule

    \multicolumn{17}{l}{\textit{1.7B Backbone}} \\
    \rowcolor{gray!15}
    Full & Qwen3 & 14.05 & 23.99 & 40.64 & 17.17 & 14.29 & 7.91 & 25.73 & 21.95 & 7.91 & 71.50 & 84.58 & 41.97 & 44.62 & 37.66 & 32.43 \\

    \midrule
    \multirow{6}{*}{Lightning}
    & Uniform & 16.79 & \cellcolor{cyan!25}17.56 & 29.21 & 22.92 & 21.97 & 9.84 & 18.53 & 21.82 & 20.47 & \cellcolor{cyan!10}66.00 & 66.36 & \cellcolor{cyan!25}40.70 & 34.58 & 40.60 & 30.53 \\
    & PostNAS & 15.19 & 15.52 & \cellcolor{cyan!10}31.36 & \cellcolor{cyan!10}29.72 & \cellcolor{cyan!25}25.67 & 10.78 & 18.48 & 22.03 & 19.25 & 61.50 & 65.41 & 39.27 & 33.25 & 40.37 & 30.56 \\
    & KL-LS & 17.38 & \cellcolor{cyan!10}16.28 & 26.54 & 22.00 & 23.47 & 8.81 & \cellcolor{cyan!10}19.07 & 21.66 & \cellcolor{cyan!25}22.61 & 61.00 & 68.24 & 38.91 & \cellcolor{cyan!25}38.97 & \cellcolor{cyan!10}45.87 & 30.77 \\
    & HALO & \cellcolor{cyan!10}18.29 & 15.43 & 27.20 & 29.23 & 23.89 & 11.15 & 18.36 & 22.07 & 20.47 & 61.50 & \cellcolor{cyan!10}72.16 & \cellcolor{cyan!10}40.09 & 37.67 & 42.58 & \cellcolor{cyan!10}31.44 \\
    & FlashMorph & 16.43 & 15.07 & 29.25 & 29.05 & 20.66 & \cellcolor{cyan!10}13.48 & 17.01 & \cellcolor{cyan!25}22.55 & 20.51 & 63.00 & \cellcolor{cyan!25}75.49 & 39.19 & 36.43 & 37.75 & 31.13 \\
    & \ourmethod & \cellcolor{cyan!25}18.98 & 15.87 & \cellcolor{cyan!25}33.28 & \cellcolor{cyan!25}31.73 & \cellcolor{cyan!10}25.29 & \cellcolor{cyan!25}13.65 & \cellcolor{cyan!25}20.62 & \cellcolor{cyan!10}22.13 & \cellcolor{cyan!10}21.15 & \cellcolor{cyan!25}67.50 & 72.12 & 39.54 & \cellcolor{cyan!10}38.60 & \cellcolor{cyan!25}46.47 & \cellcolor{cyan!25}33.35 \\

    \midrule
    \multirow{6}{*}{GLA}
    & Uniform & 14.59 & 18.63 & \cellcolor{cyan!10}29.78 & 26.08 & 20.53 & 8.99 & 15.71 & \cellcolor{cyan!25}19.62 & \cellcolor{cyan!10}20.86 & \cellcolor{cyan!25}68.50 & 66.54 & 37.65 & 39.12 & 37.87 & 30.32 \\
    & PostNAS & \cellcolor{cyan!10}16.50 & \cellcolor{cyan!25}25.50 & \cellcolor{cyan!25}30.69 & \cellcolor{cyan!10}26.51 & \cellcolor{cyan!25}24.95 & \cellcolor{cyan!25}13.83 & \cellcolor{cyan!10}18.98 & \cellcolor{cyan!10}13.97 & 18.68 & 65.50 & 65.56 & 39.16 & 37.51 & 40.04 & \cellcolor{cyan!10}31.24 \\
    & KL-LS & 15.80 & 19.36 & 24.58 & 25.69 & 24.08 & 11.12 & \cellcolor{cyan!25}19.58 & 13.84 & 19.99 & 63.00 & 69.03 & \cellcolor{cyan!25}40.32 & \cellcolor{cyan!25}43.77 & 43.61 & 30.98 \\
    & HALO & 15.00 & 23.16 & 28.97 & 20.45 & 23.26 & 9.84 & 15.63 & 12.23 & 19.92 & 66.00 & 66.79 & 35.79 & 38.09 & 43.08 & 29.87 \\
    & FlashMorph & 16.28 & 21.88 & 27.63 & 23.04 & 23.23 & \cellcolor{cyan!10}11.22 & 16.68 & 8.89 & \cellcolor{cyan!25}21.74 & 66.50 & \cellcolor{cyan!25}73.01 & 37.16 & \cellcolor{cyan!10}42.78 & \cellcolor{cyan!10}44.86 & 31.06 \\
    & \ourmethod & \cellcolor{cyan!25}16.91 & \cellcolor{cyan!10}23.84 & 29.39 & \cellcolor{cyan!25}26.80 & \cellcolor{cyan!10}24.49 & 10.22 & 18.91 & 13.17 & 15.44 & \cellcolor{cyan!10}68.00 & \cellcolor{cyan!10}72.55 & \cellcolor{cyan!10}40.05 & 42.21 & \cellcolor{cyan!25}45.37 & \cellcolor{cyan!25}31.95 \\

    \midrule
    \multirow{6}{*}{GDN}
    & Uniform & 18.45 & 23.29 & 16.85 & 28.07 & \cellcolor{cyan!25}25.92 & 12.82 & 18.96 & \cellcolor{cyan!10}22.06 & 20.42 & \cellcolor{cyan!25}70.50 & 67.77 & 39.43 & 34.82 & 35.99 & 31.10 \\
    & PostNAS & 18.63 & 25.57 & \cellcolor{cyan!10}31.52 & 26.05 & 21.63 & \cellcolor{cyan!25}14.48 & \cellcolor{cyan!10}21.68 & 21.93 & \cellcolor{cyan!25}21.57 & \cellcolor{cyan!10}68.50 & 65.95 & \cellcolor{cyan!10}40.02 & 34.50 & 38.80 & 32.20 \\
    & KL-LS & 19.11 & 25.78 & 28.82 & 25.82 & 21.50 & 11.90 & 21.50 & 21.66 & 21.08 & \cellcolor{cyan!10}68.50 & 70.47 & 39.90 & 38.40 & \cellcolor{cyan!10}43.72 & \cellcolor{cyan!10}32.73 \\
    & HALO & \cellcolor{cyan!25}19.66 & 24.59 & 31.15 & \cellcolor{cyan!25}30.99 & \cellcolor{cyan!10}22.87 & \cellcolor{cyan!10}13.03 & 20.08 & 18.68 & 16.90 & 65.50 & 66.68 & \cellcolor{cyan!25}40.19 & \cellcolor{cyan!10}39.07 & 42.80 & 32.30 \\
    & FlashMorph & 18.02 & \cellcolor{cyan!25}27.09 & 28.95 & \cellcolor{cyan!10}30.89 & 22.78 & 13.02 & 20.06 & \cellcolor{cyan!25}22.16 & 20.13 & 64.50 & \cellcolor{cyan!25}76.95 & 37.08 & 34.65 & 40.39 & 32.62 \\
    & \ourmethod & \cellcolor{cyan!10}19.32 & \cellcolor{cyan!10}26.05 & \cellcolor{cyan!25}31.79 & 27.60 & 21.71 & 12.27 & \cellcolor{cyan!25}22.02 & 21.05 & \cellcolor{cyan!10}21.09 & \cellcolor{cyan!10}68.50 & \cellcolor{cyan!10}76.51 & 38.47 & \cellcolor{cyan!25}40.30 & \cellcolor{cyan!25}46.66 & \cellcolor{cyan!25}33.81 \\

    \midrule
    \midrule
    \multicolumn{17}{l}{\textit{8B Backbone}} \\
    \rowcolor{gray!15} 
    Full & Qwen3 & 24.29 & 43.66 & 48.42 & 41.25 & 33.59 & 14.02 & 29.05 & 19.43 & 22.98 & 71.00 & 89.82 & 44.30 & 67.55 & 61.76 & 43.65 \\
    \midrule
    \multirow{4}{*}{Lightning}
    & Uniform & 20.91 & \cellcolor{cyan!25}31.88 & \cellcolor{cyan!10}40.29 & 38.04 & 27.73 & 17.10 & 22.85 & 23.52 & 20.81 & 65.00 & 69.51 & 41.17 & \cellcolor{cyan!10}44.64 & \cellcolor{cyan!25}43.51 & 36.21 \\
    & HALO & 17.08 & \cellcolor{cyan!10}30.97 & 28.25 & 37.02 & 27.51 & 16.86 & 23.17 & 22.42 & 19.27 & 60.50 & 62.56 & 38.95 & 42.18 & 41.99 & 33.48 \\
    & FlashMorph & \cellcolor{cyan!10}22.78 & 27.69 & 33.77 & \cellcolor{cyan!25}40.54 & \cellcolor{cyan!10}31.24 & \cellcolor{cyan!10}17.79 & \cellcolor{cyan!10}26.08 & \cellcolor{cyan!10}23.91 & \cellcolor{cyan!10}21.75 & \cellcolor{cyan!10}68.50 & \cellcolor{cyan!10}75.09 & \cellcolor{cyan!10}42.45 & 43.11 & 41.83 & \cellcolor{cyan!10}36.89 \\
    & \ourmethod & \cellcolor{cyan!25}27.13 & 30.63 & \cellcolor{cyan!25}40.56 & \cellcolor{cyan!10}39.10 & \cellcolor{cyan!25}35.67 & \cellcolor{cyan!25}20.11 & \cellcolor{cyan!25}27.30 & \cellcolor{cyan!25}24.20 & \cellcolor{cyan!25}22.91 & \cellcolor{cyan!25}70.50 & \cellcolor{cyan!25}77.82 & \cellcolor{cyan!25}43.68 & \cellcolor{cyan!25}49.05 & \cellcolor{cyan!10}42.34 & \cellcolor{cyan!25}39.36 \\
    \midrule
    \midrule
    \multicolumn{17}{l}{\textit{30B-A3B Backbone}} \\
    \rowcolor{gray!15} 
    Full & Qwen3 & 21.14 & 42.91 & 31.69 & 17.18 & 18.17 & 6.85 & 29.92 & 18.65 & 22.54 & 73.00 & 87.55 & 44.16 & 71.17 & 68.06 & 39.50 \\
    \midrule
    \multirow{4}{*}{Lightning}
    & Uniform & 16.50 & 30.88 & 27.93 & \cellcolor{cyan!25}40.40 & \cellcolor{cyan!10}32.85 & \cellcolor{cyan!25}22.35 & 21.66 & 21.61 & \cellcolor{cyan!10}17.30 & 60.50 & 56.33 & \cellcolor{cyan!10}41.43 & 45.50 & 45.32 & 34.33 \\
    & HALO & 17.87 & 21.02 & 26.58 & 37.43 & 29.30 & \cellcolor{cyan!10}15.89 & \cellcolor{cyan!25}26.74 & \cellcolor{cyan!10}21.73 & 13.91 & 47.50 & 68.24 & 39.01 & 47.06 & 50.06 & 33.02 \\
    & FlashMorph & \cellcolor{cyan!10}20.07 & \cellcolor{cyan!10}32.30 & \cellcolor{cyan!25}29.90 & \cellcolor{cyan!10}37.81 & 28.58 & 15.20 & \cellcolor{cyan!10}25.20 & 21.06 & 13.43 & \cellcolor{cyan!25}65.50 & \cellcolor{cyan!10}69.76 & 40.02 & \cellcolor{cyan!10}50.33 & \cellcolor{cyan!10}50.53 & \cellcolor{cyan!10}35.69 \\
    & \ourmethod & \cellcolor{cyan!25}21.43 & \cellcolor{cyan!25}33.71 & \cellcolor{cyan!10}28.90 & 37.76 & \cellcolor{cyan!25}33.03 & 13.01 & 20.65 & \cellcolor{cyan!25}21.99 & \cellcolor{cyan!25}17.81 & \cellcolor{cyan!10}62.50 & \cellcolor{cyan!25}78.09 & \cellcolor{cyan!25}42.10 & \cellcolor{cyan!25}59.42 & \cellcolor{cyan!25}55.96 & \cellcolor{cyan!25}37.60 \\
    \bottomrule
    \end{tabular}
\end{table*}

\begin{table*}[t]
    \centering
    \caption{\textbf{Performance on Language Modeling and Zero-Shot Commonsense Reasoning.} Avg denotes the average accuracy across all eight zero-shot commonsense reasoning tasks. For each backbone and linear-attention variant, the \colorbox{cyan!25}{best} and \colorbox{cyan!10}{second-best} results among layer-selection methods are highlighted.}
    \label{tab:zero_shot}
    \renewcommand{\arraystretch}{0.92}
    \resizebox{\textwidth}{!}{
    \begin{tabular}{ll|cc|cccccccc|c}
    \toprule
    \multirow{2}{*}{\textbf{Attn.}}
    & \multirow{2}{*}{\textbf{Method}} 
    & \textbf{Wiki.} & \textbf{LMB.} & \textbf{LMB.} & \textbf{ARC-e} & \textbf{ARC-c} & \textbf{PIQA} & \textbf{Hella.} & \textbf{Wino.} & \textbf{SIQA} & \textbf{BoolQ} 
    & \multirow{2}{*}{\textbf{Avg}} \\
    & & ppl $\downarrow$ & ppl $\downarrow$ & acc $\uparrow$ & acc $\uparrow$  & acc$_n$ $\uparrow$ & acc $\uparrow$  & acc$_n$ $\uparrow$ & acc $\uparrow$ & acc $\uparrow$ & acc $\uparrow$ & \\
    \midrule
    
    \multicolumn{13}{l}{\textit{0.6B Backbone}} \\
    \rowcolor{gray!15}
    Full & Qwen3 & 26.18 & 24.72 & 41.02 & 60.65 & 33.96 & 67.90 & 47.24 & 55.96 & 39.05 & 64.40 & 51.27 \\

    \midrule
    \multirow{5}{*}{Lightning}
    & Uniform & \cellcolor{cyan!10}22.58 & 12.94 & 49.18 & 65.19 & 33.79 & 68.44 & 49.01 & \cellcolor{cyan!10}57.85 & 40.79 & 61.41 & 53.21 \\
    & KL-LS & 31.43 & \cellcolor{cyan!25}12.44 & \cellcolor{cyan!10}50.01 & 64.81 & \cellcolor{cyan!25}34.64 & 68.28 & 48.75 & \cellcolor{cyan!25}58.25 & \cellcolor{cyan!25}41.25 & 60.73 & 53.34 \\
    & HALO & 23.02 & 12.74 & 49.62 & 65.91 & 34.47 & 68.28 & 48.99 & 56.35 & 40.99 & \cellcolor{cyan!25}63.94 & \cellcolor{cyan!10}53.57 \\
    & FlashMorph & 24.40 & 12.56 & 49.87 & \cellcolor{cyan!25}66.46 & 34.39 & \cellcolor{cyan!25}68.93 & \cellcolor{cyan!25}49.26 & 56.12 & 41.15 & 61.16 & 53.42 \\
    & \ourmethod & \cellcolor{cyan!25}20.20 & \cellcolor{cyan!10}12.52 & \cellcolor{cyan!25}50.32 & \cellcolor{cyan!10}65.99 & \cellcolor{cyan!10}34.50 & \cellcolor{cyan!10}68.88 & \cellcolor{cyan!10}49.15 & 57.46 & \cellcolor{cyan!10}41.20 & \cellcolor{cyan!10}61.74 & \cellcolor{cyan!25}53.66 \\

    \midrule
    \multirow{5}{*}{GLA}
    & Uniform & 21.34 & 12.53 & 50.67 & 65.78 & 33.70 & \cellcolor{cyan!25}69.86 & 49.19 & 55.88 & 41.35 & 63.03 & 53.68 \\
    & KL-LS & 29.25 & \cellcolor{cyan!25}11.93 & \cellcolor{cyan!25}51.19 & \cellcolor{cyan!10}65.95 & \cellcolor{cyan!25}36.01 & 69.26 & \cellcolor{cyan!10}49.36 & 55.80 & \cellcolor{cyan!25}41.61 & \cellcolor{cyan!10}65.11 & \cellcolor{cyan!25}54.29 \\
    & HALO & 27.29 & \cellcolor{cyan!10}12.15 & \cellcolor{cyan!10}51.14 & \cellcolor{cyan!25}66.04 & \cellcolor{cyan!10}34.47 & 68.72 & 49.20 & \cellcolor{cyan!10}56.04 & 41.30 & \cellcolor{cyan!25}66.67 & \cellcolor{cyan!10}54.20 \\
    & FlashMorph & \cellcolor{cyan!10}20.41 & 12.41 & 50.61 & 65.74 & 33.96 & 68.93 & 49.34 & 54.93 & 41.20 & 64.59 & 53.66 \\
    & \ourmethod & \cellcolor{cyan!25}20.14 & 12.37 & 50.15 & 64.94 & 34.39 & \cellcolor{cyan!10}69.37 & \cellcolor{cyan!25}49.52 & \cellcolor{cyan!25}56.27 & \cellcolor{cyan!10}41.50 & 63.46 & 53.70 \\

    \midrule
    \multirow{5}{*}{GDN}
    & Uniform & 20.17 & 12.48 & 50.24 & 65.78 & 34.13 & \cellcolor{cyan!25}68.99 & 49.43 & \cellcolor{cyan!10}55.72 & 41.20 & 63.49 & 53.63 \\
    & KL-LS & 19.83 & 12.16 & 50.65 & 65.74 & 35.32 & 68.88 & \cellcolor{cyan!10}49.64 & \cellcolor{cyan!25}56.67 & \cellcolor{cyan!25}41.81 & 64.25 & 54.12 \\
    & HALO & 20.05 & 12.30 & 50.69 & 65.95 & \cellcolor{cyan!10}35.58 & 68.72 & 49.27 & 55.17 & 41.10 & 64.40 & 53.86 \\
    & FlashMorph & \cellcolor{cyan!25}19.64 & \cellcolor{cyan!25}11.87 & \cellcolor{cyan!25}51.33 & \cellcolor{cyan!25}66.46 & \cellcolor{cyan!25}35.92 & \cellcolor{cyan!10}68.93 & 49.56 & 55.25 & 41.61 & \cellcolor{cyan!10}65.38 & \cellcolor{cyan!25}54.31 \\
    & \ourmethod & \cellcolor{cyan!10}19.72 & \cellcolor{cyan!10}12.03 & \cellcolor{cyan!10}51.12 & \cellcolor{cyan!10}66.08 & 34.47 & \cellcolor{cyan!25}68.99 & \cellcolor{cyan!25}49.80 & 55.25 & \cellcolor{cyan!10}41.66 & \cellcolor{cyan!25}65.78 & \cellcolor{cyan!10}54.14 \\

    \midrule
    \midrule

    \multicolumn{13}{l}{\textit{1.7B Backbone}} \\
    \rowcolor{gray!15}
    Full & Qwen3 & 20.98 & 12.19 & 51.41 & 72.22 & 42.66 & 72.31 & 60.36 & 61.56 & 45.04 & 77.65 & 60.40 \\

    \midrule
    \multirow{6}{*}{Lightning}
    & Uniform & 15.26 & 6.91 & \cellcolor{cyan!10}59.64 & \cellcolor{cyan!10}76.73 & 44.45 & 74.16 & 62.16 & 62.19 & 44.01 & \cellcolor{cyan!25}77.86 & 62.65 \\
    & PostNAS & \cellcolor{cyan!10}14.79 & \cellcolor{cyan!10}6.85 & \cellcolor{cyan!25}59.65 & \cellcolor{cyan!25}77.53 & 44.54 & 74.32 & \cellcolor{cyan!10}62.33 & \cellcolor{cyan!10}63.93 & \cellcolor{cyan!25}44.98 & 77.22 & \cellcolor{cyan!25}63.06 \\
    & KL-LS & 14.87 & 6.93 & 59.34 & 75.42 & \cellcolor{cyan!25}46.93 & 74.32 & 62.16 & 62.27 & \cellcolor{cyan!10}44.63 & 73.91 & 62.37 \\
    & HALO & 14.91 & 7.05 & 58.72 & 76.47 & 44.97 & \cellcolor{cyan!25}74.81 & 62.03 & 61.01 & 43.86 & 77.46 & 62.42 \\
    & FlashMorph & 14.84 & 6.98 & 59.07 & 76.35 & 45.22 & 74.27 & 62.30 & \cellcolor{cyan!25}64.09 & 44.27 & 77.52 & 62.89 \\
    & \ourmethod & \cellcolor{cyan!25}14.71 & \cellcolor{cyan!25}6.77 & 59.54 & 76.64 & \cellcolor{cyan!10}45.65 & \cellcolor{cyan!10}74.43 & \cellcolor{cyan!25}62.51 & 62.98 & 43.91 & \cellcolor{cyan!10}77.55 & \cellcolor{cyan!10}62.90 \\

    \midrule
    \multirow{6}{*}{GLA}
    & Uniform & 14.91 & 6.72 & \cellcolor{cyan!10}61.17 & \cellcolor{cyan!25}77.61 & 45.82 & 74.27 & 62.70 & 62.83 & 44.88 & \cellcolor{cyan!10}77.37 & \cellcolor{cyan!10}63.33 \\
    & PostNAS & \cellcolor{cyan!10}14.69 & 6.68 & 60.28 & 76.81 & 45.22 & 74.05 & 62.67 & \cellcolor{cyan!25}64.09 & 44.63 & 76.97 & 63.09 \\
    & KL-LS & 14.70 & \cellcolor{cyan!25}6.63 & 60.61 & \cellcolor{cyan!10}77.23 & \cellcolor{cyan!25}47.35 & 74.37 & \cellcolor{cyan!25}62.75 & 62.43 & \cellcolor{cyan!10}45.09 & 75.57 & 63.18 \\
    & HALO & 14.92 & 6.68 & \cellcolor{cyan!25}62.56 & 76.98 & 45.99 & 74.32 & 62.56 & \cellcolor{cyan!10}62.98 & 44.47 & 75.55 & 63.18 \\
    & FlashMorph & 14.73 & \cellcolor{cyan!10}6.64 & 60.59 & 76.89 & 45.73 & \cellcolor{cyan!25}74.65 & \cellcolor{cyan!10}62.71 & 62.59 & 44.78 & \cellcolor{cyan!25}78.17 & 63.26 \\
    & \ourmethod & \cellcolor{cyan!25}14.60 & \cellcolor{cyan!10}6.64 & 60.47 & 76.77 & \cellcolor{cyan!10}46.93 & \cellcolor{cyan!10}74.48 & \cellcolor{cyan!10}62.71 & \cellcolor{cyan!25}64.09 & \cellcolor{cyan!25}45.29 & 76.91 & \cellcolor{cyan!25}63.46 \\

    \midrule
    \multirow{6}{*}{GDN}
    & Uniform & 14.65 & 6.81 & \cellcolor{cyan!25}60.37 & 75.60 & 45.56 & 74.48 & 62.75 & 62.19 & 44.01 & \cellcolor{cyan!10}77.80 & 62.85 \\
    & PostNAS & 14.45 & 6.74 & 59.98 & \cellcolor{cyan!25}77.15 & \cellcolor{cyan!25}46.33 & 74.16 & 62.54 & \cellcolor{cyan!25}64.33 & 44.42 & 77.68 & \cellcolor{cyan!25}63.32 \\
    & KL-LS & \cellcolor{cyan!10}14.35 & \cellcolor{cyan!10}6.68 & 60.06 & 76.60 & 45.56 & \cellcolor{cyan!25}74.86 & \cellcolor{cyan!25}63.00 & 62.75 & \cellcolor{cyan!25}44.78 & 76.94 & 63.07 \\
    & HALO & 14.46 & 6.77 & 59.67 & 76.60 & \cellcolor{cyan!10}45.99 & \cellcolor{cyan!10}74.54 & 62.69 & \cellcolor{cyan!10}63.93 & \cellcolor{cyan!10}44.52 & 77.43 & \cellcolor{cyan!10}63.17 \\
    & FlashMorph & 14.40 & 6.78 & 59.83 & 76.85 & 45.05 & 74.16 & \cellcolor{cyan!10}62.84 & 63.14 & 43.76 & \cellcolor{cyan!25}78.10 & 62.97 \\
    & \ourmethod & \cellcolor{cyan!25}14.33 & \cellcolor{cyan!25}6.67 & \cellcolor{cyan!10}60.26 & \cellcolor{cyan!10}76.89 & 45.05 & 74.32 & 62.82 & 62.75 & \cellcolor{cyan!10}44.52 & 77.71 & 63.04 \\

    \midrule
    \midrule
    \multicolumn{13}{l}{\textit{8B Backbone}} \\
    \rowcolor{gray!15} 
    Full & Qwen3 & 12.20 & 4.60 & 65.09 & 83.38 & 56.48 & 76.61 & 74.88 & 67.48 & 51.69 & 86.61 & 70.28 \\
    \midrule
    \multirow{4}{*}{Lightning}
    & Uniform & 11.15 & 4.27 & 68.95 & \cellcolor{cyan!10}81.99 & 55.46 & \cellcolor{cyan!10}78.45 & 74.22 & 71.59 & 49.08 & \cellcolor{cyan!10}82.20 & 70.24 \\
    & HALO & 11.27 & \cellcolor{cyan!10}4.20 & \cellcolor{cyan!10}68.99 & 81.86 & \cellcolor{cyan!10}55.97 & \cellcolor{cyan!25}78.89 & 74.18 & 71.59 & 48.82 & 81.56 & 70.23 \\
    & FlashMorph & \cellcolor{cyan!10}11.09 & 4.26 & 68.91 & 81.65 & \cellcolor{cyan!25}56.06 & \cellcolor{cyan!25}78.89 & \cellcolor{cyan!10}74.43 & \cellcolor{cyan!10}71.74 & \cellcolor{cyan!25}49.59 & \cellcolor{cyan!25}83.03 & \cellcolor{cyan!10}70.54 \\
    & \ourmethod & \cellcolor{cyan!25}10.87 & \cellcolor{cyan!25}4.15 & \cellcolor{cyan!25}70.35 & \cellcolor{cyan!25}82.03 & 55.72 & 78.24 & \cellcolor{cyan!25}74.70 & \cellcolor{cyan!25}73.01 & \cellcolor{cyan!10}49.13 & 82.11 & \cellcolor{cyan!25}70.66 \\
    \midrule
    \midrule
    \multicolumn{13}{l}{\textit{30B-A3B Backbone}} \\
    \rowcolor{gray!15} 
    Full & Qwen3 & 10.89 & 4.13 & 65.63 & 79.29 & 55.72 & 79.43 & 77.69 & 70.09 & 49.49 & 88.75 & 70.76 \\
    \midrule
    \multirow{4}{*}{Lightning}
    & Uniform & \cellcolor{cyan!10}10.20 & 3.77 & 71.78 & 82.28 & 57.42 & 79.60 & \cellcolor{cyan!10}77.02 & 72.77 & \cellcolor{cyan!10}48.93 & \cellcolor{cyan!25}84.34 & 71.77 \\
    & HALO & \cellcolor{cyan!25}10.11 & \cellcolor{cyan!25}3.67 & \cellcolor{cyan!10}72.40 & \cellcolor{cyan!10}82.95 & \cellcolor{cyan!25}58.45 & 79.49 & 76.82 & \cellcolor{cyan!10}73.80 & 48.46 & \cellcolor{cyan!10}83.79 & \cellcolor{cyan!25}72.02 \\
    & FlashMorph & 10.28 & 3.77 & 71.84 & 82.91 & \cellcolor{cyan!10}57.59 & \cellcolor{cyan!10}79.65 & 76.78 & \cellcolor{cyan!25}74.03 & 48.31 & 83.61 & 71.84 \\
    & \ourmethod & \cellcolor{cyan!25}10.11 & \cellcolor{cyan!10}3.75 & \cellcolor{cyan!25}72.48 & \cellcolor{cyan!25}83.21 & 57.42 & \cellcolor{cyan!25}79.76 & \cellcolor{cyan!25}77.03 & 73.56 & \cellcolor{cyan!25}48.98 & 83.24 & \cellcolor{cyan!10}71.96 \\
    \bottomrule
    \end{tabular}
    }
\end{table*}


\end{document}